%% file: main.tex
\documentclass[sigconf,nonacm]{acmart}
\usepackage{array}

\setcopyright{none}
\acmDOI{}
\acmISBN{}

\AtBeginDocument{\raggedbottom}

\title{When to Intervene? An Empirical Study of Errors and Self-Recovery in LLM Agents}
\author{Chiming Duan}
\email{duanchiming@stu.pku.edu.cn}
\affiliation{
  \institution{Peking University}
  \city{Beijing}
  \country{China}
}

\author{Tong Jia}
\email{jia.tong@pku.edu.cn}
\affiliation{
  \institution{Peking University}
  \city{Beijing}
  \country{China}
}

\author{Yingxi Chen}
\email{chenyingxi@bupt.edu.cn}
\affiliation{
  \institution{Beijing University of Posts and Telecommunications}
  \city{Beijing}
  \country{China}
}

\author{Minghua He}
\email{hemh2120@stu.pku.edu.cn}
\affiliation{
  \institution{Peking University}
  \city{Beijing}
  \country{China}
}

\author{Lingzhe Zhang}
\email{zhang.lingzhe@stu.pku.edu.cn}
\affiliation{
  \institution{Peking University}
  \city{Beijing}
  \country{China}
}

\author{Ying Li}
\email{li.ying@pku.edu.cn}
\affiliation{
  \institution{Peking University}
  \city{Beijing}
  \country{China}
}

\renewcommand{\shortauthors}{Duan et al.}
\hypersetup{pdftitle={When to Intervene? An Empirical Study of Errors and Self-Recovery in LLM Agents},pdfauthor={Chiming Duan, Tong Jia, Yingxi Chen, Minghua He, Lingzhe Zhang, Ying Li}}

\begin{document}
\begin{abstract}
LLM agents interact with Web services to perform tasks on users' behalf. Errors during these interactions can disrupt execution and prevent task completion, motivating external intervention. However, agents may self-recover through continued execution, raising the question of when to intervene and when to wait for self-recovery. We present an empirical study of errors and self-recovery in LLM agents. We construct ReInBench, a dataset containing 1,259 agent traces and 1,403 error events across computer-use and Web tasks. Our study examines how often and how quickly agents self-recover, how self-recovery relates to subsequent execution and task completion, and whether self-recovery can be predicted when an error occurs. We find that 725 error events self-recover, often within a few steps, with self-recovery performance differing across Execution LLMs. Self-recovery can also help agents recover from subsequent errors of the same type and support overall task completion. Intervention should therefore prioritize errors unlikely to self-recover, while allowing agents to continue when self-recovery is likely. Building on these empirical findings, we propose ReCoF, a method that predicts whether an error will self-recover to inform whether additional intervention is needed. ReCoF achieves an average relative improvement of 9.02\% in Macro-F1 over the strongest baseline on ReInBench.
Our code and data can be found at \url{https://method-notes.github.io/ReCoF/}.
\end{abstract}
\maketitle

\section{Introduction}
\label{sec:introduction}

LLM agents provide a means for users to delegate tasks to Web services through natural-language instructions \citep{zhou2024webarena}. They mediate users' interactions with Web services by translating goals into actions and interpreting feedback \citep{cai2025personalized}. For example, computer-use agents (CUAs) can operate websites through graphical browser interfaces. Because an agent determines the actions performed on a user's behalf, its errors can result in incorrect requests to Web applications or incomplete task workflows \citep{shi2025toolwild}. Error detection is therefore important for enabling timely intervention so that agents' Web interactions satisfy users' goals.

Existing error detection methods inspect action validity and unfolding agent traces. These methods examine the agent's execution context to identify errors or anticipate failures \citep{barbi2025preventing,zhang2026agentforesight}. When a problem is identified, they trigger an intervention, such as requesting a replacement action or injecting corrective guidance into the agent's context \citep{ji2026sentry,nanda2026wink}. Such interventions aim to redirect execution and restore task progress. However, an agent may also self-recover through continued execution, raising the question of when external intervention is needed and when the agent should be allowed to self-recover.

Some studies examine when external intervention is beneficial by comparing execution with and without observation refresh \citep{xiao2026harnesses}. These methods estimate when a specific recovery operation improves task outcomes, but do not explicitly characterize the agent's self-recovery process or its role in subsequent execution. External intervention may interrupt this process before the agent has an opportunity to self-recover. Self-recovery may also influence how agents handle later errors within the same task. Reflective feedback from unsuccessful attempts can improve performance in later trials \citep{shinn2023reflexion}, whereas stored experiences can also propagate earlier errors \citep{xiong2026memory}. However, how naturally occurring self-recovery relates to subsequent error handling and overall task completion remains insufficiently understood. Understanding when to intervene therefore requires examining both the agent's ability to self-recover and the consequences of its self-recovery process.

To investigate these issues, we construct ReInBench, a dataset containing 1,259 agent traces and 1,403 error events, of which 725 self-recover within the observed execution. It includes computer-use and Web tasks covering browsing, document and spreadsheet editing, and workflows spanning multiple applications. ReInBench records error occurrence, recognition when observable, and recovery outcomes, together with the corresponding execution context. These records support analysis of self-recovery frequency and speed, its relationship with subsequent error handling and task completion, and self-recovery prediction to inform intervention decisions. We address three research questions.

\begin{description}
    \item[RQ1] How often and how quickly do agents self-recover from errors, and how does this vary across LLMs?
    \item[RQ2] Does self-recovery help agents recover from subsequent errors and complete their tasks?
    \item[RQ3] When an error occurs, should we intervene or wait for self-recovery?
\end{description}

Together, these questions examine how agents self-recover, how recovery relates to subsequent execution, and how these findings can inform intervention decisions. Building on the results of these RQs, we design ReCoF, a method that predicts whether an error will self-recover to inform whether additional intervention is needed. On the one hand, ReCoF trains \emph{Recovery Pattern Experts} on historical execution contexts and recovery outcomes to capture patterns in how agents understand and handle errors. On the other hand, it uses \emph{Conditional Decision Fusion} to adjust expert contributions according to differences in recovery capability and the difficulty of the current error. Our study makes three contributions.
\begin{itemize}
    \item We construct \textbf{ReInBench}, a dataset containing 1,259 agent traces and 1,403 error events across computer-use and Web tasks. It records error occurrence, observable recognition, recovery outcomes, and execution context, supporting the study of self-recovery and intervention decisions.
    \item We conduct an empirical study of \textbf{errors and self-recovery in LLM agents}, examining self-recovery frequency, speed, and variation across Execution LLMs. We further analyze how earlier self-recovery relates to subsequent recovery of same-type and different-type errors, and how local recovery relates to overall task completion.
    \item We propose \textbf{ReCoF}, a method that predicts whether an error will self-recover to inform whether additional intervention is needed. ReCoF achieves an average relative improvement of 9.02\% in Macro-F1 over the strongest baseline on ReInBench.
\end{itemize}


\section{Background}
\label{sec:background}

\begin{figure}[htbp]
    \centering
    \includegraphics[width=\columnwidth]{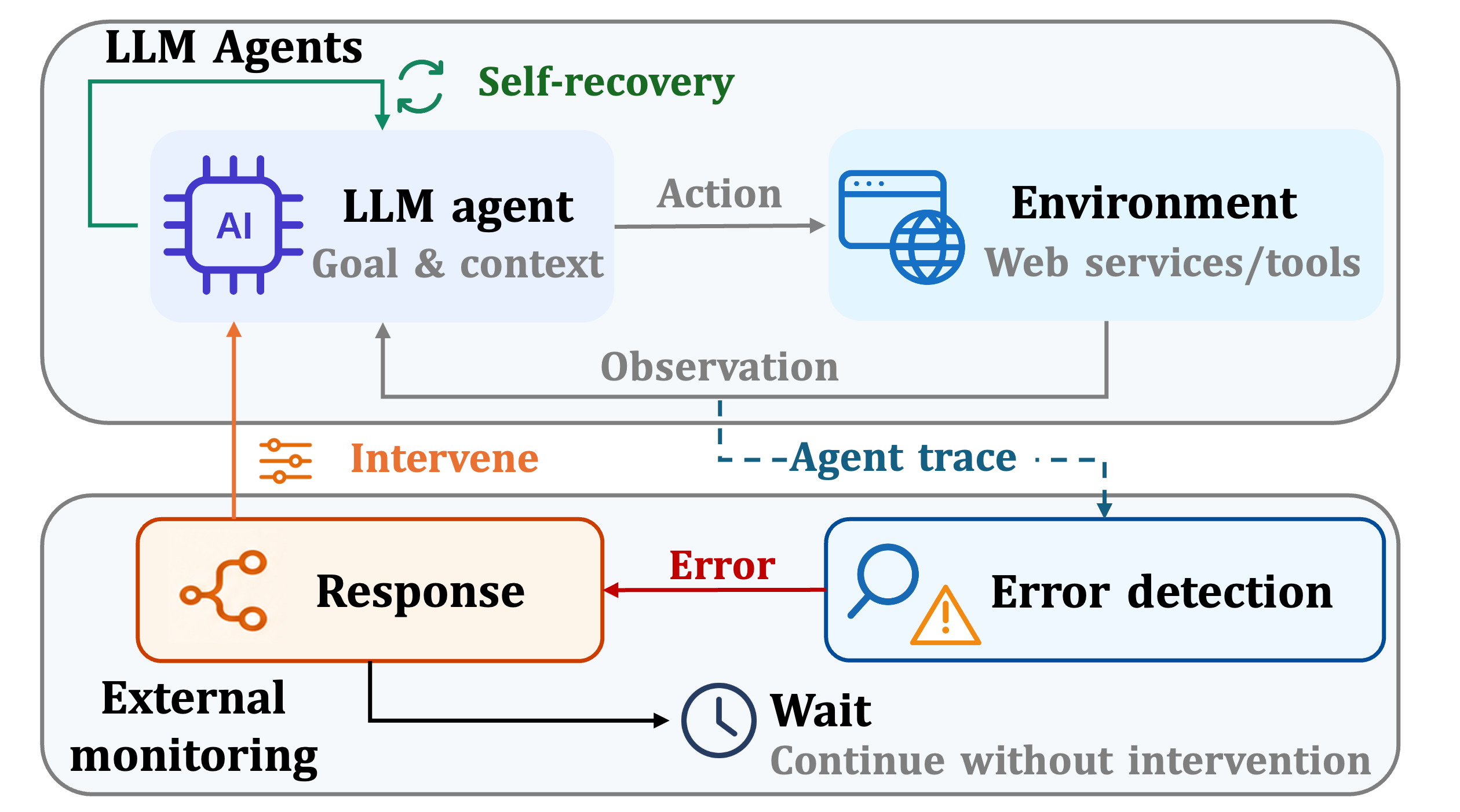}
    \caption{Agent self-recovery and error detection.}
    \Description{An LLM agent exchanges actions and feedback with an environment. An internal loop represents self-recovery. An external monitor receives the current trace and detects errors. A separate response can add guidance to the agent or wait without adding guidance.}
    \label{fig:agent-error-recovery}
\end{figure}

\subsection{LLM Agents}
\label{sec:background-agents}

An LLM agent is a system that uses an LLM for reasoning and planning, with tools or interfaces that execute actions toward a user-specified goal. In Web applications, agents connect users' instructions to operations on Web services \citep{cai2025personalized}. These operations can be exposed through callable functions or graphical interfaces. Tool-using agents invoke functions with generated arguments \citep{shi2025toolwild}, whereas computer-use agents interact through mouse and keyboard actions \citep{xie2024osworld}. Both forms of interaction produce feedback that informs later decisions. ReAct describes this interaction as interleaved reasoning and acting, in which observations help the agent update its plans and handle exceptions \citep{yao2023react}.

During this interaction, the agent produces reasoning outputs and actions and receives observations from the environment. The time-ordered record of these steps forms an \emph{agent trace}, showing how the agent interprets feedback and acts toward the task goal. Figure~\ref{fig:agent-error-recovery} illustrates the execution loop. Let $g$ denote the task goal, $a_t$ the action at step $t$, $o_t$ the observation returned after that action, and $r_t$ the agent's recorded reasoning, if available. With $o_0$ denoting the initial observation, the \emph{current trace} contains the execution history available after step $t$.
\begin{equation}
    \tau_{\leq t} = \bigl(o_0, (r_i,a_i,o_i)_{i=1}^{t}\bigr).
    \label{eq:current-trace}
\end{equation}
For an execution of $T$ steps, the full agent trace is $\tau=\tau_{\leq T}$. The agent selects its next action according to $a_{t+1}\sim\pi(\cdot\mid g,\tau_{\leq t})$, where $\pi$ denotes the agent's policy. Observations include screenshots, page content, or tool responses, depending on the interaction interface.

\subsection{Errors and Self-Recovery}
\label{sec:background-recovery}

\paragraph{Errors.}
An \emph{error} is an incorrect system state that may lead to \emph{failure}, a deviation from the required service \citep{avizienis2004basic}. In LLM agents, errors include misinterpreted task requirements, incorrect tool calls \citep{shi2025toolwild}, and unsupported assumptions or repeated actions that stall progress \citep{ji2026sentry}. An error detector $D$ evaluates execution against the task goal $g$ using the current trace $\tau_{\leq t}$ (Eq.~\eqref{eq:current-trace}) \citep{zhang2026agentforesight}.
\begin{equation}
    \hat e_t = D(g,\tau_{\leq t}) \in \{0,1\}.
    \label{eq:error-detection}
\end{equation}
Here, $\hat e_t=1$ flags an error at step $t$, and $\hat e_t=0$ indicates that no error is detected.

\begin{figure}[t]
    \centering
    \includegraphics[width=\columnwidth]{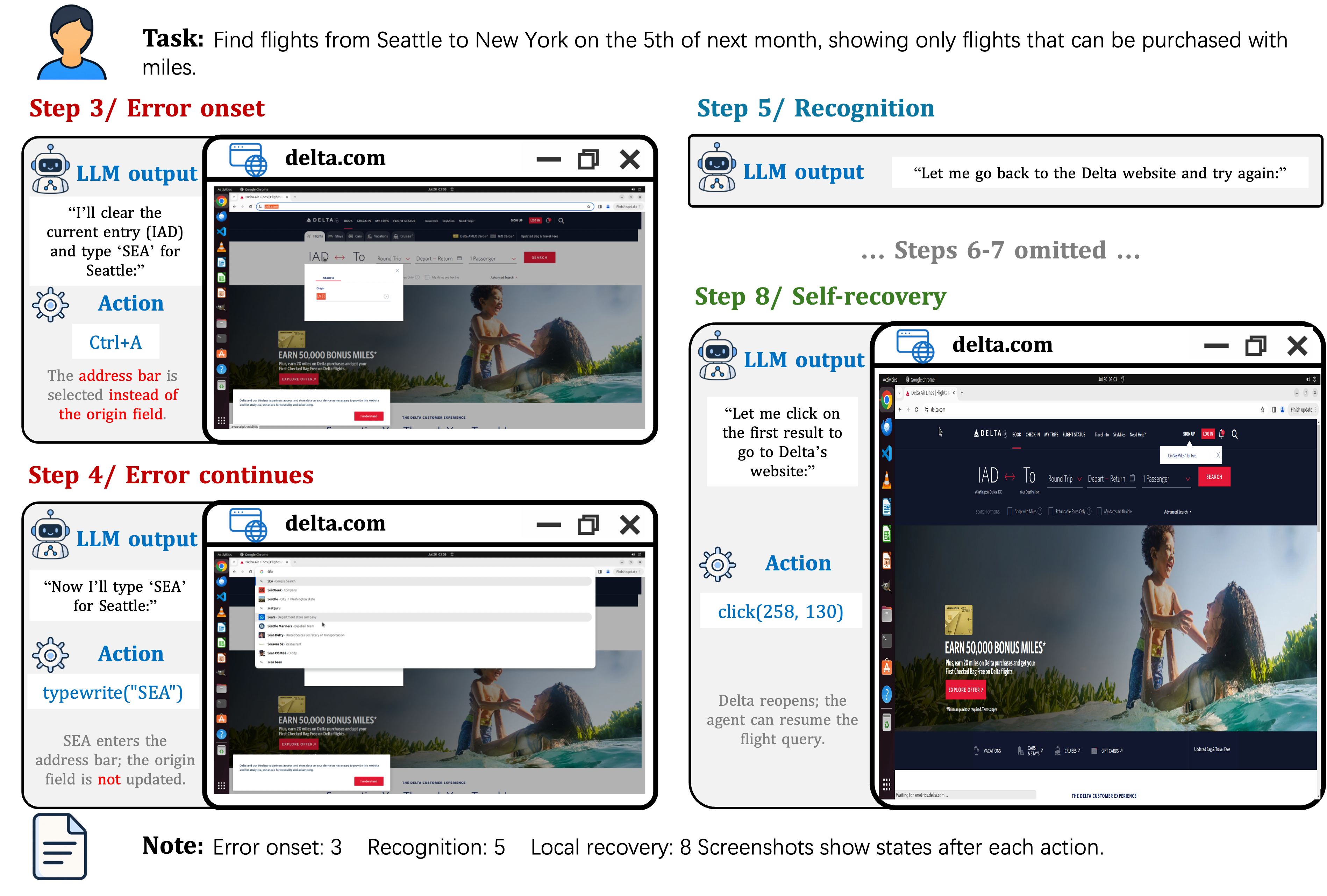}
    \caption{A Web self-recovery example.}
    \Description{A two-column montage of a Delta flight-query trace. The left column shows step 3 error onset and step 4 continued error, with LLM output, actions and screenshots. The right column shows the agent's recognition at step 5, omitted steps 6 and 7, and local self-recovery at step 8. Ctrl+A selects the address bar, SEA is entered there instead of the origin field, and the agent later reopens Delta to continue querying.}
    \label{fig:web-error-case}
\end{figure}

Figure~\ref{fig:web-error-case} illustrates error onset, recognition, and self-recovery in a Web task trace from ReInBench \citep{osworldverifiedtraces}. At step 3, the agent selects the address bar instead of the departure field. It types ``SEA'' there at step 4. It recognizes the problem at step 5 and reopens Delta at step 8 to resume the query.

\paragraph{Self-recovery.}
\emph{Self-recovery} is the process through which an agent resolves an error using its own subsequent reasoning and actions, without additional external assistance. It follows the original agent policy $\pi$, including its built-in retry and reflection mechanisms \citep{yao2023react,wu2025guireflection}, and concerns resolution of the local error rather than completion of the entire task. We characterize this process through error onset, recognition, and recovery. Recognition is observable when the agent explicitly acknowledges an error or exhibits targeted corrective behavior. However, some errors reflect temporary deviations in reasoning or action selection and can be resolved as ordinary task execution restores progress. For illustration, suppose an agent has established that one product exceeds the budget and another meets all requirements, but continues switching between their pages without acquiring new information. It subsequently uses the available evidence to select the qualifying product, ending the unproductive loop. Here, continued task reasoning resolves the error without revealing a separate diagnosis of the redundant browsing. Recovery is therefore observable, while a distinct recognition point may not be identifiable. We therefore annotate self-recovery and observable recognition separately.

\paragraph{Intervention timing.}
Some existing methods respond to detected failures or misbehaviors with corrective guidance. For example, Sentry retrieves failure-specific lessons, while Wink provides targeted course corrections \citep{ji2026sentry,nanda2026wink}. When self-recovery is possible, the timing of this assistance becomes a separate question. In this study, we represent the choice after an error flag $\hat e_t=1$ with a response policy $C$.
\begin{equation}
    d_t=C(g,\tau_{\leq t})\in\{\textsc{Wait},\textsc{Intervene}\}.
    \label{eq:intervention-response}
\end{equation}
\emph{Intervening} supplies external corrective guidance. \emph{Waiting} allows $\pi$ to continue, with the decision revisited as $\tau_{\leq t}$ grows. This timing choice weighs the opportunity for self-recovery against the delay and cost of further execution.

\section{ReInBench}
\label{sec:methodology}
\label{sec:reinbench}

ReInBench contains 1,259 agent traces and 1,403 error events from agents performing Web browsing, document and spreadsheet editing, and cross-application workflows. Among these events, 725 self-recover: 553 have an observable recognition point and 172 recover without one. It records error onset, observable recognition, and self-recovery outcomes to support the study of intervention timing. We describe data collection in Section~\ref{sec:data-collection}, annotation in Section~\ref{sec:annotation}, and experimental settings in Section~\ref{sec:data-analysis}.

\input{sections/methodology-rq1}

\section{Results}
\label{sec:results}

\subsection{Self-Recovery Performance (RQ1)}
\label{sec:rq1}
\input{sections/rq1}

\input{sections/rq3-results}

\subsection{Subsequent Execution (RQ2)}
\label{sec:rq2}
\input{sections/rq2}

\subsection{Intervention Timing (RQ3)}
\label{sec:rq3}
\input{sections/rq3}

\subsection{Threats to Validity}
\label{sec:threats}
\textbf{Dataset coverage.} ReInBench covers a limited set of Execution LLMs, task environments, and error types. Self-recovery behavior may differ across applications, task complexity, and model capabilities, so the observed patterns may not generalize to all agent settings. Broader coverage of models, tasks, and error scenarios is needed to assess the generality of our findings. \textbf{When and how to intervene.} Our study focuses on \emph{when} to intervene by predicting whether an error will self-recover. However, the effectiveness of intervention also depends on \emph{how} assistance is provided, including its content and compatibility with the agent's current execution. RQ3 evaluates self-recovery prediction without executing interventions. Further experiments are needed to examine how intervention timing and intervention methods jointly affect task completion.

\section{Related Work}
\label{sec:related-work}

\input{sections/related-work}

\section{Conclusion}
\label{sec:conclusion}
In this paper, we presented an empirical study of errors and self-recovery in LLM agents to investigate when external intervention is needed. We constructed ReInBench to analyze self-recovery across computer-use and Web tasks. Our results show that agents self-recover from a substantial share of errors, often within a few steps, with recovery performance differing across Execution LLMs. Self-recovery can also help agents handle subsequent errors of the same type and support overall task completion. These findings motivate allowing opportunities for self-recovery and prioritizing intervention for errors likely to remain unresolved. Building on the empirical findings, we proposed ReCoF, a method that predicts whether an error will self-recover to inform whether additional intervention is needed. ReCoF achieves an average relative improvement of 9.02\% in Macro-F1 over the strongest baseline on ReInBench. Our study establishes self-recovery as an important consideration in intervention decisions. Future work will broaden the coverage of ReInBench and examine how intervention timing and intervention methods jointly affect task completion.

\bibliographystyle{ACM-Reference-Format}
\bibliography{references}
\clearpage
\appendix
\input{appendices/recof-method}
\end{document}

%% file: sections/methodology-rq1.tex
\subsection{Data Collection}
\label{sec:data-collection}
We study LLM agents performing Web browsing, desktop application tasks, and cross-application workflows. The tasks include interacting with Web services, editing documents and spreadsheets, and coordinating actions across applications. We construct ReInBench from agent traces collected from two source datasets. OSWorld-Verified \citep{osworldverifiedtraces,xie2024osworld} contains agent traces generated by computer-use agents (CUAs) built on Claude 3.7 Sonnet, Claude Sonnet 4.5, and Kimi K2.5. These agents perform tasks in browser and desktop applications. For Online-Mind2Web \citep{xue2025online,deng2023mind2web}, we use the HAL dataset of agent traces generated by SeeAct with GPT-5-2025-08-07 at medium reasoning effort \citep{hal2025seeact}. We refer to the corresponding task groups in ReInBench as OSWorld tasks and Web tasks, respectively. Table~\ref{tab:study-datasets} summarizes the source datasets and Execution LLMs represented in ReInBench.

\begin{table}[htbp]
\caption{ReInBench overview.}
\label{tab:study-datasets}
\centering\footnotesize
\setlength{\tabcolsep}{3pt}
\renewcommand{\arraystretch}{1.08}
\begin{tabular*}{\columnwidth}{@{}l@{\extracolsep{\fill}}rrr@{}}
\toprule
\textbf{Execution LLM} & \textbf{Traces} & \textbf{Errors} & \textbf{Recovered} \\
\midrule
\multicolumn{4}{@{}l@{}}{\textbf{OSWorld-Verified}\quad 358 tasks} \\
\addlinespace[3pt]
Claude 3.7 Sonnet & 304 & 588 & 268 \\
Claude Sonnet 4.5 & 339 & 343 & 226 \\
Kimi K2.5 & 343 & 271 & 165 \\
\midrule
\multicolumn{4}{@{}l@{}}{\textbf{Online-Mind2Web}\quad 273 tasks} \\
\addlinespace[3pt]
GPT-5 (2025-08-07) & 273 & 201 & 66 \\
\midrule
Total & 1,259 & 1,403 & 725 \\
\bottomrule
\end{tabular*}
\end{table}

\paragraph{OSWorld-Verified.}
We select 1,076 agent traces covering 360 tasks for annotation. The sample includes 360 from CUAs using Claude 3.7 Sonnet, 360 using Claude Sonnet 4.5, and 356 using Kimi K2.5. After checking annotation completeness and event consistency, we retain 986 agent traces covering 358 tasks. These agents operate through graphical computer-use interfaces. Each agent trace pairs agent outputs and actions with screenshots, enabling inspection of the environment before and after corrective actions.

\paragraph{Online-Mind2Web.}
We check 300 agent traces for field completeness and unique step alignment and select 291 agent traces covering 291 tasks for annotation. Annotation completes for 273 agent traces covering 273 tasks. These agent traces were generated by SeeAct with GPT-5-2025-08-07 at medium reasoning effort on August 13, 2025. Each agent trace records the task instruction, agent outputs, action feedback, and selected-element metadata. The dataset does not include original screenshots or complete DOM snapshots.

\subsection{Annotation}
\label{sec:annotation}
\paragraph{Annotation unit and labels.}
We annotate error events in ReInBench traces, with multiple events allowed per trace. Each event records its onset, the agent's recognition point, and self-recovery point when observable. We also specify a self-recovery criterion that describes the observable state establishing that the error or its task-relevant obstruction has been removed. Onset is the first recorded occurrence of the error. Recognition is the earliest step showing that the agent itself identifies the error through an explicit acknowledgment or targeted corrective behavior. Self-recovery is recorded at the earliest step whose observed state satisfies the criterion. An effective rollback or valid alternative path can establish self-recovery. Self-recovery does not require an observable recognition point: an error may be resolved during the agent's continued execution without an explicit acknowledgment or identifiable targeted correction. Each error event is labeled as recovered or not recovered within the observed trace. A recurrence after verified local self-recovery starts a new event. Repeated failed attempts before self-recovery remain in the same event. For errors involving failed action execution in Web tasks, subsequent action feedback can establish self-recovery when it confirms that the intended action succeeds. A criterion concerning a page or field state requires later feedback or element metadata supporting that state. We establish self-recovery from observable execution evidence that satisfies the event's recovery criterion.

\paragraph{Annotation procedure.}
We use LLM-assisted annotation followed by expert verification, drawing on prior work on human--LLM collaborative annotation and uncertainty-guided allocation of annotation effort \citep{wang2024collaborative,li2023coannotating}. GPT-6.1-sol and GLM-5.3-Flash independently identify and annotate error events under shared guidelines without access to each other's annotations. Each annotation records the event boundaries, self-recovery outcome and criterion, and references to supporting agent trace records. A verification stage combines rule-based checks with LLM-based evidence review. The two annotations are compared to identify unmatched events and conflicting labels or boundaries. Rule-based checks validate annotation structure, step ranges, temporal consistency, and references to source records. LLM-based review examines the supporting evidence and identifies judgments requiring closer inspection. Disagreements between the annotation LLMs and verifier flags guide the order and depth of expert review. Three researchers studying LLM agents form the expert panel, comprising two doctoral students and one master's student. All three participate in developing the annotation guidelines. Every model-generated annotation undergoes manual verification. Each panel member independently checks whether the error occurs, its onset, the agent's recognition point, the self-recovery outcome and boundary when observable, and the self-recovery criterion and supporting evidence. Reviewers can see the model labels and explanations but not the other members' judgments. When the two annotation LLMs disagree, the panel re-examines the entire agent trace and jointly re-annotates the events through discussion until consensus is reached. This re-annotation reassesses event identification, labels, and temporal boundaries. Ambiguities found during manual verification receive the same full-trace re-annotation. Final decisions follow the annotation guidelines and observable execution evidence. Initial model outputs, verification results, and adjudication records are preserved to make label changes traceable.

\subsection{Experimental Settings}
\label{sec:data-analysis}
\begingroup
\emergencystretch=2em
\paragraph{For RQ1,} we examine self-recovery frequency, recognition and self-recovery timing, and differences across Execution LLMs in ReInBench. First, we investigate how often agents resolve errors through self-recovery. For each LLM, we calculate the proportion of error events that self-recover within the observed agent trace. Second, we investigate how quickly agents self-recover and where delays arise in this process. We measure onset-to-recovery intervals for 600 recovered events in OSWorld tasks and 66 in Web tasks, with identifiable onset and recovery boundaries. As discussed in Section~\ref{sec:background-recovery}, self-recovery can occur without a separately observable recognition point. To examine potential bottlenecks in recognition and correction, we therefore use the 509 events in OSWorld tasks and 44 in Web tasks with an observable recognition point between onset and recovery, and compare onset-to-recognition and recognition-to-recovery intervals. Events without an identifiable recognition point contribute only to the end-to-end timing analysis. Third, we investigate how self-recovery varies across Execution LLMs performing the same tasks. We compare self-recovery proportions and timing across the three LLMs on shared tasks. For tasks in which all three LLMs encounter errors, we further compare whether each LLM resolves all of its errors within the observed trace.\par
\endgroup

\paragraph{For RQ2,} we investigate whether prior self-recovery helps agents recover from subsequent errors and complete their tasks. First, to investigate whether agents benefit from earlier self-recovery, we construct ordered error-event pairs $(e_i,e_j)$ within each trace $\tau$ in ReInBench, where $e_i$ has self-recovered before $e_j$ begins. Each recovered event is paired with every subsequent event satisfying this condition. This procedure yields 613 error-event pairs, including 200 same-type pairs and 413 different-type pairs. We annotate each pair as containing either same-type or different-type errors. For example, two errors in which the agent types into the wrong input field form a same-type pair. A wrong-field input error paired with an incorrect file-path error forms a different-type pair. Within these pairs, we examine whether $e_j$ self-recovers to assess whether the recovery experience from $e_i$ may help the agent resolve the later error. When both events recover and their temporal boundaries are identifiable, we compare the number of steps from error onset to recovery to assess whether the later error is resolved faster. We first analyze these patterns across all pairs and then group them according to whether $e_i$ and $e_j$ are the same type of error, to investigate whether subsequent recovery outcomes and speed vary with error similarity. Second, to investigate whether self-recovery supports overall task completion, we analyze recovery at the trace level. For each error-bearing trace, we calculate the proportion of annotated errors that self-recover. Using the recorded task evaluation outcomes, we divide traces into successful and unsuccessful groups and compare their mean self-recovery proportions. We further group traces into those with no recovery, partial recovery, or complete recovery, and compare task completion rates across these groups to examine how the extent of local self-recovery relates to overall task completion.

\paragraph{For RQ3,} we investigate when to intervene after an error occurs. An agent may self-recover through continued execution, allowing the task to proceed without external assistance. Immediate intervention may therefore be unnecessary, whereas waiting for an error that remains unresolved can delay task progress. To assess whether intervention is needed, we predict, at the time an error occurs, whether the agent will subsequently self-recover, formulating this prediction as a binary classification task. A predicted self-recovery supports allowing execution to continue, while predicted non-recovery indicates a potential need for assistance. We use the recovery outcomes recorded in ReInBench as target labels, with non-recovery indicating that the error remains unresolved within the observed trace. Inputs include the task goal and up to five recent execution steps ending at the error occurrence, covering recorded reasoning, actions, feedback, and available screenshots. Future execution and recovery annotations are excluded. We propose ReCoF, a method that predicts whether an error will self-recover to inform whether additional intervention is needed. Figure~\ref{fig:recof-overview} illustrates its workflow. On the one hand, ReCoF trains \emph{Recovery Pattern Experts} on historical execution contexts and their self-recovery outcomes, learning predictive recovery patterns from text, visual, and joint image--text representations. On the other hand, ReCoF uses an Assessment LLM to examine the task requirements and the Execution LLM's observed reasoning and actions, assessing how difficult the current error may be for the agent to recover from and providing a contextual verdict. The \emph{Conditional Decision Fusion} module then uses this verdict to adaptively weight the expert predictions and produce the final self-recovery prediction. Appendix~\ref{app:recof-method} provides the formal definitions and training procedure. We conduct this evaluation on CUA tasks in ReInBench, using task-disjoint training, validation, and test partitions. We compare methods that can be used to predict self-recovery, including Direct LLM, AgentForesight, Sentry, PrefixGuard, and Wink \citep{zhang2026agentforesight,ji2026sentry,huang2026prefixguard,nanda2026wink}. These comparisons assess whether the execution context available when an error occurs can support self-recovery prediction and inform the choice between waiting and intervention.

\input{sections/rq3-method-figure}

\paragraph{Evaluation metrics.}
For RQ1, we measure self-recovery frequency as $\rho=R/(R+N)$, where $R$ and $N$ denote recovered and unrecovered error events. Self-recovery speed is measured by onset-to-recovery, onset-to-recognition, and recognition-to-recovery intervals, summarized through their distributions and medians. For RQ2, we report the proportion of later errors that self-recover and compare recovery durations within event pairs. A later recovery is faster or equal when $\Delta_j\leq\Delta_i$, where $\Delta_i$ and $\Delta_j$ denote the earlier and later recovery durations. At the trace level, we compare mean self-recovery proportions and task completion rates, defined as the proportion of traces that successfully complete their tasks. For RQ3, we report precision, recall, F1, Macro-F1, and accuracy, treating non-recovery as the positive class. Precision and recall are $\mathrm{Prec}=TP/(TP+FP)$ and $\mathrm{Rec}=TP/(TP+FN)$, with $F_1=2\,\mathrm{Prec}\,\mathrm{Rec}/(\mathrm{Prec}+\mathrm{Rec})$. Macro-F1 gives equal weight to both classes, calculated as $\mathrm{Macro}\text{-}F_1=(F_{1,\mathrm{recovery}}+F_{1,\mathrm{non\text{-}recovery}})/2$.

\paragraph{Comparison methods.}
For RQ3, we compare ReCoF with the following methods, adapting existing monitoring approaches to self-recovery prediction.
\begin{itemize}
    \item \textbf{Direct LLM} directly predicts whether an error will self-recover from the task goal and available execution context.
    \item \textbf{AgentForesight} \citep{zhang2026agentforesight} identifies decisive errors from the execution context. We adapt its forecasting procedure to distinguish errors that will self-recover from those that will remain unresolved.
    \item \textbf{Sentry} \citep{ji2026sentry} monitors execution using knowledge learned from previous failures. We adapt its failure taxonomy to guide self-recovery prediction from the task goal and execution context.
    \item \textbf{PrefixGuard} \citep{huang2026prefixguard} learns event representations and uses a recurrent predictor to estimate failure risk from the execution context. We adapt its prediction target to whether the current error will self-recover.
    \item \textbf{Wink} \citep{nanda2026wink} identifies specification drift, reasoning problems, and tool call failures to guide corrective intervention. We adapt its misbehavior analysis to predict self-recovery without executing the intervention.
\end{itemize}
To examine the contributions of ReCoF's components, we evaluate two ablation variants.
\begin{itemize}
    \item \textbf{w/o Recovery Pattern Experts} removes the expert predictions and their conditional interactions, retaining the Assessment LLM's contextual verdict and Execution LLM identity.
    \item \textbf{w/o Conditional Interaction} removes only the interaction terms, retaining both the expert predictions and contextual verdict, together with Execution LLM identity.
\end{itemize}
Each variant is refitted and calibrated using the validation set. We further examine parameter sensitivity by separately varying the fusion regularization parameter $C$, the number of task-grouped folds $K$, and the offset $\Delta$ from the calibrated decision threshold, while holding the remaining configuration fixed.

\paragraph{Implementation details.}
For RQ3, we use GLM-5.3-Flash as ReCoF's Assessment LLM and for the LLM-based baselines, with temperature set to zero. These methods receive the task goal and up to five steps of execution context ending at the error occurrence, including execution text and available screenshots. We use task-disjoint partitions containing 559 training, 228 validation, and 238 test events. ReCoF uses frozen text, visual, and joint image--text encoders to train its Recovery Pattern Experts. Its logistic fusion uses $C=0.01$ and three task-grouped folds on the validation set. We retain the three fold models, average their test probabilities, and calibrate a separate decision threshold for each Execution LLM using validation predictions. PrefixGuard uses 4,096-dimensional TF-IDF features and a GRU predictor, trained for 24 epochs with AdamW at a learning rate of $10^{-3}$. Its checkpoint and thresholds are selected using validation Macro-F1. We report results over five seeds for PrefixGuard, ReCoF, and its ablations, keeping ReCoF's Recovery Pattern Experts and Assessment LLM outputs fixed across fusion runs.

%% file: sections/rq3-method-figure.tex
\begin{figure}[t]
\centering
\includegraphics[width=\columnwidth]{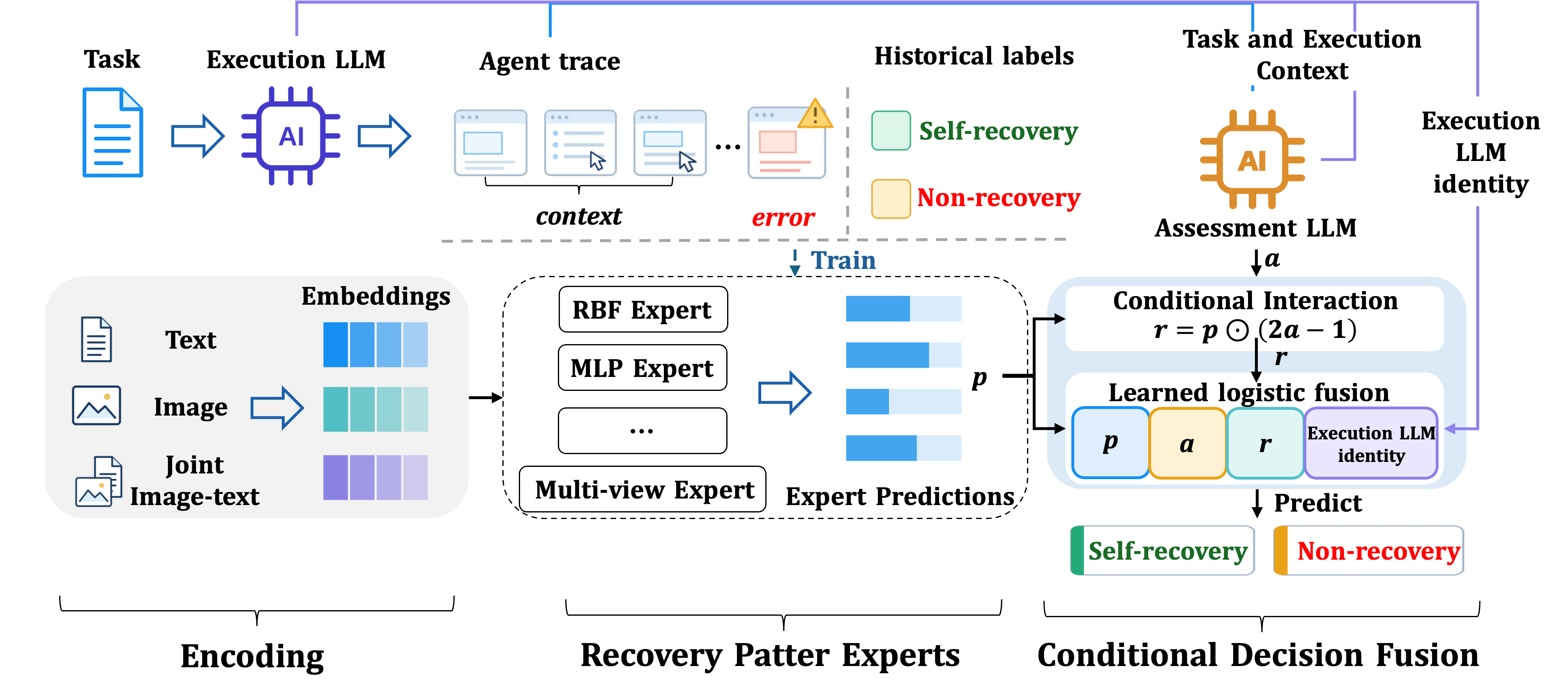}
\caption{ReCoF overview.}
\Description{Overview of ReCoF. Text, image, and joint image-text embeddings feed Recovery Pattern Experts that produce predictions p. An Assessment LLM provides a contextual verdict a. Conditional interaction computes r from p and a, and learned logistic fusion combines p, a, r, and Execution LLM identity to predict self-recovery or non-recovery. Historical execution contexts and recovery outcomes supply training supervision.}
\label{fig:recof-overview}
\end{figure}

%% file: sections/rq1.tex
Figure~\ref{fig:rq1-status} compares self-recovery proportions and task-level overlap in ReInBench. Figure~\ref{fig:rq1-timing} shows the timing of error recognition and self-recovery. We make the following observations.

\input{sections/rq1-floats}

\textbf{Agents self-recover from a substantial share of errors.} In ReInBench, 659 of 1,202 errors in OSWorld tasks self-recover within the observed agent trace, yielding a proportion of 54.8\%. For Web tasks, 66 of 201 errors self-recover, yielding 32.8\%. These results show that agents can resolve many local errors through continued execution without additional external assistance. Self-recovery is therefore an important consideration when deciding whether to intervene after an error occurs.

\textbf{Self-recovery is generally fast, but correction can require prolonged execution.} Figure~\ref{fig:rq1-timing}a shows that end-to-end recovery durations concentrate within the first few steps, with medians of two steps for OSWorld tasks and three steps for Web tasks. Figures~\ref{fig:rq1-timing}b,c further show that onset-to-recognition intervals are concentrated around one step, whereas recognition-to-recovery intervals are more dispersed. For OSWorld tasks, 73.3\% of events with observable recognition recover within two subsequent steps. However, the correction distributions extend beyond the initial peak, particularly for Web tasks, indicating that recognizing an error does not always lead to an immediate resolution. The end-to-end distributions also exhibit long tails, with recovery durations reaching 78 steps for OSWorld tasks and 48 steps for Web tasks. These results suggest that a short continuation allows many errors to resolve, while some require substantially more execution before recovery.

\textbf{Execution LLMs differ in both self-recovery frequency and task coverage.} Figure~\ref{fig:rq1-status}a shows self-recovery proportions on ReInBench's OSWorld tasks of 45.6\% for Claude 3.7 Sonnet, 65.9\% for Claude Sonnet 4.5, and 60.9\% for Kimi K2.5. On shared tasks with errors in all three LLMs, the proportions are 43.8\%, 64.4\%, and 52.7\%, respectively. Sonnet 4.5 exceeds Claude 3.7 by 20.6 percentage points on these shared tasks, so its higher self-recovery proportion within the Claude family persists on the same task set. Figure~\ref{fig:rq1-overlap}b further compares task-level recovery across these LLMs. A task belongs to an LLM's set when all of its annotated errors recover. We normalize percentages over shared tasks meeting this condition for at least one LLM. Within this pool, 7.8\% satisfy the condition for all three LLMs, whereas 52.9\% do so for only one. Claude Sonnet 4.5 and Kimi K2.5 share a pair-only region of 25.5\%. The overlap shows that the tasks on which all annotated errors self-recover differ across Execution LLMs. Across the broader shared-task set, each LLM has a median observed recovery duration of two steps, indicating similar typical recovery speed despite differences in recovery frequency and task coverage.

%% file: sections/rq1-floats.tex
\begin{figure}[t]
\centering
\includegraphics[width=\columnwidth]{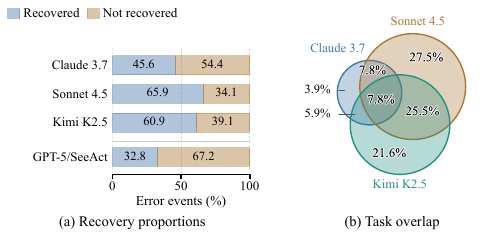}
\caption{Self-recovery across Execution LLMs.}
\Description{Two side-by-side panels. Panel (a) shows shares of error events that recovered or did not recover within the observed trace. Recovery shares in OSWorld tasks are 45.6 percent for Claude 3.7 Sonnet, 65.9 percent for Claude Sonnet 4.5, and 60.9 percent for Kimi K2.5. The separate GPT-5/SeeAct row for Web tasks has 32.8 percent recovered. Panel (b) shows task-level overlap on shared OSWorld tasks with errors in all three LLMs. Each set contains tasks where that LLM recovers all annotated errors. Percentages use the union of these sets as 100 percent. Exclusive shares are 3.9, 27.5, and 21.6 percent. Pair-only shares are 7.8, 5.9, and 25.5 percent, and the three-way share is 7.8 percent. Circle areas are approximate.}
\label{fig:rq1-status}
\label{fig:rq1-overlap}
\end{figure}

\begin{figure*}[t]
\centering
\includegraphics[width=\textwidth]{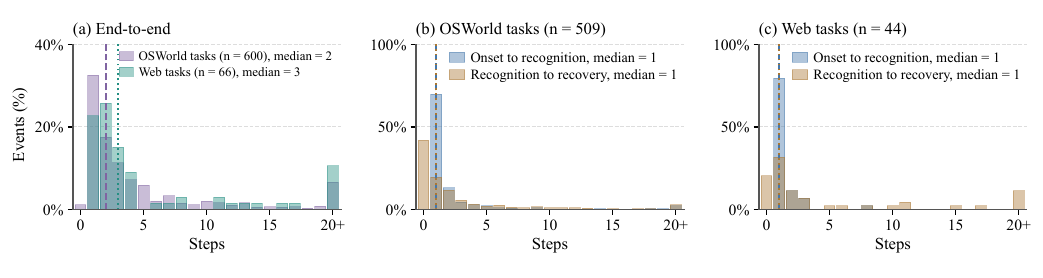}
\caption{Self-recovery timing.}
\Description{Three side-by-side percentage-frequency histograms with median markers. Panel (a) compares end-to-end onset-to-recovery duration for 600 events in OSWorld tasks and 66 in Web tasks, including recoveries without observable recognition. Panel (b) compares onset-to-recognition and recognition-to-recovery intervals within 509 events in OSWorld tasks. Panel (c) compares these two stages within 44 events in Web tasks. Each series is normalized separately, and the final bin contains intervals of 20 or more steps. Steps use each task group's recorded execution units in ReInBench.}
\label{fig:rq1-timing}
\end{figure*}

%% file: sections/rq3-results.tex
\begin{table*}[t]
\caption{Self-recovery prediction and core ablations on ReInBench.}
\label{tab:rq3-results}
\centering
\begingroup
\fontsize{7}{8.2}\selectfont
\setlength{\tabcolsep}{1.6pt}
\renewcommand{\arraystretch}{1.12}
\begin{tabular*}{\textwidth}{@{}l@{\extracolsep{\fill}}*{15}{r}@{}}
\toprule
\textbf{Method / variant} & \multicolumn{5}{c}{\raisebox{-1.4pt}{\includegraphics[height=8pt]{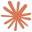}}\hspace{3pt}\textbf{Claude 3.7 Sonnet}} & \multicolumn{5}{c}{\raisebox{-1.4pt}{\includegraphics[height=8pt]{figures/model-logos/claude-icon.png}}\hspace{3pt}\textbf{Claude Sonnet 4.5}} & \multicolumn{5}{c}{\raisebox{-1.4pt}{\includegraphics[height=8pt]{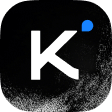}}\hspace{3pt}\textbf{Kimi K2.5}} \\
\cmidrule(lr){2-6}\cmidrule(lr){7-11}\cmidrule(l){12-16}
 & \textbf{M-F1} & \textbf{F1} & \textbf{Prec.} & \textbf{Rec.} & \textbf{Acc.} & \textbf{M-F1} & \textbf{F1} & \textbf{Prec.} & \textbf{Rec.} & \textbf{Acc.} & \textbf{M-F1} & \textbf{F1} & \textbf{Prec.} & \textbf{Rec.} & \textbf{Acc.} \\
\midrule
Direct LLM & $68.75$ & $62.50$ & $69.77$ & $56.60$ & $70.00$ & $72.88$ & $63.41$ & $72.22$ & $56.52$ & $76.19$ & $61.69$ & $47.06$ & $50.00$ & $44.44$ & $67.27$ \\
AgentForesight & $61.49$ & $47.37$ & $\textcolor[HTML]{B51F24}{\mathbf{78.26}}$ & $33.96$ & $66.67$ & $68.89$ & $55.56$ & $\textcolor[HTML]{B51F24}{\mathbf{76.92}}$ & $43.48$ & $74.60$ & $47.78$ & $17.39$ & $40.00$ & $11.11$ & $65.45$ \\
Sentry & $61.10$ & $48.78$ & $68.97$ & $37.74$ & $65.00$ & $72.88$ & $63.41$ & $72.22$ & $56.52$ & $76.19$ & $50.82$ & $24.00$ & $42.86$ & $16.67$ & $65.45$ \\
PrefixGuard & \begin{tabular}[c]{@{}r@{}}$58.59$\\[-2pt]{\tiny$\pm 3.19$}\end{tabular} & \begin{tabular}[c]{@{}r@{}}$59.40$\\[-2pt]{\tiny$\pm 1.77$}\end{tabular} & \begin{tabular}[c]{@{}r@{}}$52.96$\\[-2pt]{\tiny$\pm 3.40$}\end{tabular} & \begin{tabular}[c]{@{}r@{}}$68.30$\\[-2pt]{\tiny$\pm 6.86$}\end{tabular} & \begin{tabular}[c]{@{}r@{}}$58.83$\\[-2pt]{\tiny$\pm 3.15$}\end{tabular} & \begin{tabular}[c]{@{}r@{}}$47.50$\\[-2pt]{\tiny$\pm 6.74$}\end{tabular} & \begin{tabular}[c]{@{}r@{}}$22.34$\\[-2pt]{\tiny$\pm 8.39$}\end{tabular} & \begin{tabular}[c]{@{}r@{}}$41.19$\\[-2pt]{\tiny$\pm 17.81$}\end{tabular} & \begin{tabular}[c]{@{}r@{}}$15.65$\\[-2pt]{\tiny$\pm 5.83$}\end{tabular} & \begin{tabular}[c]{@{}r@{}}$59.68$\\[-2pt]{\tiny$\pm 7.24$}\end{tabular} & \begin{tabular}[c]{@{}r@{}}$55.43$\\[-2pt]{\tiny$\pm 5.72$}\end{tabular} & \begin{tabular}[c]{@{}r@{}}$43.85$\\[-2pt]{\tiny$\pm 8.22$}\end{tabular} & \begin{tabular}[c]{@{}r@{}}$42.90$\\[-2pt]{\tiny$\pm 8.71$}\end{tabular} & \begin{tabular}[c]{@{}r@{}}$50.00$\\[-2pt]{\tiny$\pm 17.12$}\end{tabular} & \begin{tabular}[c]{@{}r@{}}$59.64$\\[-2pt]{\tiny$\pm 8.47$}\end{tabular} \\
Wink & $66.26$ & $59.79$ & $65.91$ & $54.72$ & $67.50$ & $66.09$ & $52.63$ & $66.67$ & $43.48$ & $71.43$ & $41.56$ & $18.18$ & $20.00$ & $16.67$ & $50.91$ \\
\midrule
w/o Recovery Pattern Experts & $68.75$ & $62.50$ & $69.77$ & $56.60$ & $70.00$ & $72.88$ & $63.41$ & $72.22$ & $56.52$ & $76.19$ & $61.69$ & $47.06$ & $50.00$ & $44.44$ & $67.27$ \\
w/o Conditional Interaction & \begin{tabular}[c]{@{}r@{}}$71.14$\\[-2pt]{\tiny$\pm 2.34$}\end{tabular} & \begin{tabular}[c]{@{}r@{}}$68.01$\\[-2pt]{\tiny$\pm 1.95$}\end{tabular} & \begin{tabular}[c]{@{}r@{}}$69.32$\\[-2pt]{\tiny$\pm 5.98$}\end{tabular} & \begin{tabular}[c]{@{}r@{}}$68.30$\\[-2pt]{\tiny$\pm 10.87$}\end{tabular} & \begin{tabular}[c]{@{}r@{}}$71.83$\\[-2pt]{\tiny$\pm 2.53$}\end{tabular} & \begin{tabular}[c]{@{}r@{}}$61.50$\\[-2pt]{\tiny$\pm 7.72$}\end{tabular} & \begin{tabular}[c]{@{}r@{}}$45.28$\\[-2pt]{\tiny$\pm 14.25$}\end{tabular} & \begin{tabular}[c]{@{}r@{}}$60.07$\\[-2pt]{\tiny$\pm 3.62$}\end{tabular} & \begin{tabular}[c]{@{}r@{}}$38.26$\\[-2pt]{\tiny$\pm 17.77$}\end{tabular} & \begin{tabular}[c]{@{}r@{}}$68.57$\\[-2pt]{\tiny$\pm 3.79$}\end{tabular} & \begin{tabular}[c]{@{}r@{}}$69.74$\\[-2pt]{\tiny$\pm 2.35$}\end{tabular} & \begin{tabular}[c]{@{}r@{}}$53.76$\\[-2pt]{\tiny$\pm 3.98$}\end{tabular} & \begin{tabular}[c]{@{}r@{}}$\textcolor[HTML]{B51F24}{\mathbf{87.42}}$\\[-2pt]{\tiny$\pm 1.13$}\end{tabular} & \begin{tabular}[c]{@{}r@{}}$38.89$\\[-2pt]{\tiny$\pm 3.93$}\end{tabular} & \begin{tabular}[c]{@{}r@{}}$\textcolor[HTML]{B51F24}{\mathbf{78.18}}$\\[-2pt]{\tiny$\pm 1.29$}\end{tabular} \\
\midrule
\textbf{ReCoF} & \begin{tabular}[c]{@{}r@{}}$\textcolor[HTML]{B51F24}{\mathbf{73.53}}$\\[-2pt]{\tiny$\pm 2.33$}\end{tabular} & \begin{tabular}[c]{@{}r@{}}$\textcolor[HTML]{B51F24}{\mathbf{71.61}}$\\[-2pt]{\tiny$\pm 0.31$}\end{tabular} & \begin{tabular}[c]{@{}r@{}}$69.56$\\[-2pt]{\tiny$\pm 5.62$}\end{tabular} & \begin{tabular}[c]{@{}r@{}}$\textcolor[HTML]{B51F24}{\mathbf{74.72}}$\\[-2pt]{\tiny$\pm 7.01$}\end{tabular} & \begin{tabular}[c]{@{}r@{}}$\textcolor[HTML]{B51F24}{\mathbf{73.83}}$\\[-2pt]{\tiny$\pm 2.54$}\end{tabular} & \begin{tabular}[c]{@{}r@{}}$\textcolor[HTML]{B51F24}{\mathbf{74.18}}$\\[-2pt]{\tiny$\pm 3.44$}\end{tabular} & \begin{tabular}[c]{@{}r@{}}$\textcolor[HTML]{B51F24}{\mathbf{66.83}}$\\[-2pt]{\tiny$\pm 2.04$}\end{tabular} & \begin{tabular}[c]{@{}r@{}}$72.19$\\[-2pt]{\tiny$\pm 13.27$}\end{tabular} & \begin{tabular}[c]{@{}r@{}}$\textcolor[HTML]{B51F24}{\mathbf{64.35}}$\\[-2pt]{\tiny$\pm 8.36$}\end{tabular} & \begin{tabular}[c]{@{}r@{}}$\textcolor[HTML]{B51F24}{\mathbf{76.51}}$\\[-2pt]{\tiny$\pm 4.55$}\end{tabular} & \begin{tabular}[c]{@{}r@{}}$\textcolor[HTML]{B51F24}{\mathbf{72.99}}$\\[-2pt]{\tiny$\pm 2.93$}\end{tabular} & \begin{tabular}[c]{@{}r@{}}$\textcolor[HTML]{B51F24}{\mathbf{62.74}}$\\[-2pt]{\tiny$\pm 2.93$}\end{tabular} & \begin{tabular}[c]{@{}r@{}}$71.05$\\[-2pt]{\tiny$\pm 14.67$}\end{tabular} & \begin{tabular}[c]{@{}r@{}}$\textcolor[HTML]{B51F24}{\mathbf{58.89}}$\\[-2pt]{\tiny$\pm 9.30$}\end{tabular} & \begin{tabular}[c]{@{}r@{}}$77.09$\\[-2pt]{\tiny$\pm 4.19$}\end{tabular} \\
\bottomrule
\end{tabular*}
\par\vspace{3pt}
\begin{minipage}{\textwidth}\scriptsize
Values are percentages. M-F1: Macro-F1. F1, precision and recall treat observed non-recovery as positive. 
\textcolor[HTML]{B51F24}{\textbf{Bold red values}} mark the highest mean in each Execution LLM--metric group. 
Smaller values give $\pm$ sample standard deviation over five seeds: classifier-training seeds for PrefixGuard and fusion-layer seeds for ReCoF and its ablations. 
ReCoF's Recovery Pattern Experts and Assessment LLM outputs are held fixed. API baselines use cached predictions. Ablations are refitted and calibrated on validation data.
SDs are omitted for w/o Recovery Pattern Experts because its predictions are identical across seeds.
\end{minipage}
\endgroup
\end{table*}

%% file: sections/rq2.tex
\begin{figure}[t]
\centering
\includegraphics[width=\columnwidth]{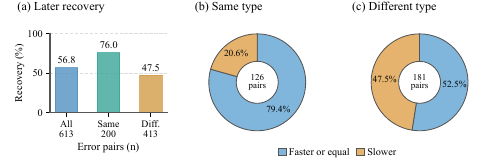}
\caption{Subsequent error recovery.}
\Description{Three side-by-side panels pooling ReInBench tasks and LLMs. Panel (a) shows later-event recovery proportions of 56.8, 76.0 and 47.5 percent among all 613, 200 same-type and 413 different-type pairs. Pairs may share a later event. Panels (b) and (c) use donut charts to compare later versus earlier recovery duration for same-type and different-type pairs, respectively. Among 126 same-type timing pairs, 79.4 percent are faster or equal and 20.6 percent slower. Among 181 different-type timing pairs, 52.5 percent are faster or equal and 47.5 percent slower. Each ring represents all valid timing pairs of its type.}
\label{fig:rq2-pairs}
\end{figure}
\begin{figure}[t]
\centering
\includegraphics[width=\columnwidth]{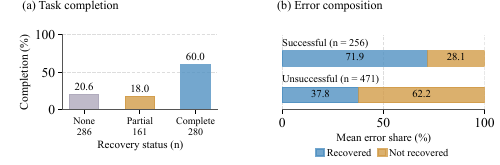}
\caption{Task completion and self-recovery.}
\Description{Two panels pooling ReInBench tasks and LLMs. Panel (a) shows task completion rates of 20.6, 18.0 and 60.0 percent for 286 traces with no recovery, 161 with partial recovery and 280 with complete recovery. Panel (b) shows the mean within-trace recovered and unrecovered error shares. The 256 successful traces average 71.9 percent recovered and 28.1 percent unrecovered. The 471 unsuccessful traces average 37.8 percent recovered and 62.2 percent unrecovered. These are averages of per-trace proportions, rather than pooled event counts.}
\label{fig:rq2-completion}
\end{figure}

Figure~\ref{fig:rq2-pairs} examines whether self-recovery from earlier errors in ReInBench helps agents recover from subsequent errors and do so more quickly. Figure~\ref{fig:rq2-completion} investigates whether self-recovery helps agents complete the overall task. We make the following observations.

\textbf{Self-recovery experience helps agents recover from subsequent errors of the same type.} Across all 613 error-event pairs, 56.8\% of later errors self-recover, while this proportion reaches 76.0\% for same-type pairs (Figure~\ref{fig:rq2-pairs}a). Recovering from an earlier error may provide useful feedback and corrective actions that the agent can reuse when encountering a similar error within the same task. The recovery-duration comparisons further support this interpretation. We compare recovery speed using the number of steps taken to recover from the earlier and later errors within each pair. For same-type pairs, 79.4\% of later recoveries take no more steps than the earlier recovery, compared with 63.5\% overall (Figure~\ref{fig:rq2-pairs}b,c). However, evidence that earlier recovery experience helps agents recover from different-type errors remains limited. For these pairs, 47.5\% of later errors self-recover, and 52.5\% of later recoveries are faster or equal in duration. Earlier recovery strategies may be less directly applicable because different-type errors can require different information or corrective actions. Different error types may also vary in difficulty and the number of steps needed for recovery, which can contribute to the observed duration differences.

\textbf{Self-recovery can support overall task completion.} Self-recovery addresses local errors, so we further examine whether its benefits extend to overall task completion. Figure~\ref{fig:rq2-completion}b shows that successful traces have a higher mean proportion of recovered errors than unsuccessful traces, at 71.9\% versus 37.8\%. Figure~\ref{fig:rq2-completion}a further shows how task completion varies with the extent of recovery. Among 727 error-bearing traces, the completion rate reaches 60.0\% when all annotated errors recover, compared with 20.6\% when none recover and 18.0\% when only some recover. Complete recovery is accompanied by a higher task completion rate, whereas partial recovery has a rate similar to no recovery. Although recovering from individual errors can restore progress, other unresolved errors may continue to obstruct execution and prevent task completion.

%% file: sections/rq3.tex
RQ3 asks when to intervene after an error occurs and when to allow the agent to self-recover. RQ1 shows that agents often self-recover within a few steps, so an error may already be resolved by the time external intervention takes effect. RQ2 further indicates that self-recovery can support subsequent error handling and overall task success. Together, these findings suggest that choosing between intervention and waiting requires predicting whether the current error will self-recover. We therefore investigate whether the execution context available when an error occurs can support this prediction and inform the intervention decision.

To investigate this question, we propose ReCoF, a method that predicts whether an error will self-recover to inform whether additional intervention is needed. Its design draws on two insights from RQ1 and RQ2. RQ1 shows that self-recovery performance differs across Execution LLMs, motivating predictions that account for the Execution LLM and its task context. RQ2 further suggests that agents' self-recovery processes reflect how they understand and handle errors, providing information useful for predicting recoverability. Accordingly, ReCoF incorporates two core components. \emph{Recovery Pattern Experts} use text, visual, and joint image--text representations of historical execution contexts, together with their recovery outcomes, to learn patterns that reflect how agents understand and handle errors. \emph{Conditional Decision Fusion} uses the Assessment LLM to assess the current error in relation to the task requirements and the Execution LLM's observed reasoning and actions. It combines this contextual verdict with the Execution LLM's identity and adaptively weights the expert predictions, accounting for differences in recovery capability across models and the difficulty of the current error.

We compare ReCoF with five baselines that can be used to predict self-recovery, including Direct LLM, AgentForesight, Sentry, PrefixGuard, and Wink. We further examine the contributions of its core components through ablation experiments and assess its sensitivity to parameter choices. Table~\ref{tab:rq3-results} and Figure~\ref{fig:rq3-sensitivity} present the results. We make the following observations.

\input{sections/rq3-sensitivity-figure}

\textbf{Execution context available when an error occurs can help distinguish self-recovery from non-recovery.} ReCoF achieves mean Macro-F1 scores of 73.53\%, 74.18\%, and 72.99\% for Claude 3.7 Sonnet, Claude Sonnet 4.5, and Kimi K2.5, respectively, exceeding the strongest baseline for each Execution LLM by 4.78, 1.30, and 11.30 percentage points. Its non-recovery recall also increases from 56.60\% to 74.72\%, from 56.52\% to 64.35\%, and from 44.44\% to 58.89\%, respectively. The improvement is smallest for Sonnet 4.5, where Direct LLM and Sentry already provide competitive predictions. These results show that information available at error occurrence supports self-recovery prediction, and combining learned recovery patterns with contextual assessment helps identify more errors that remain unresolved.

\textbf{Historical recovery patterns and contextual assessment provide complementary evidence of recoverability.} Removing the Recovery Pattern Experts reduces Macro-F1 to 68.75\%, 72.88\%, and 61.69\% across the three Execution LLMs. This variant produces the same binary predictions as Direct LLM, showing that historical recovery patterns provide useful information beyond the Assessment LLM's contextual verdict. Removing the conditional interaction while retaining both prediction sources reduces Macro-F1 to 71.14\%, 61.50\%, and 69.74\%. The benefit therefore comes from both the available prediction sources and how their contributions are combined. For Kimi, removing the interaction increases non-recovery precision to 87.42\% but lowers recall to 38.89\%, compared with 71.05\% precision and 58.89\% recall for ReCoF. Conditioning expert contributions on the contextual verdict improves the balance between precision and recall, helping identify unresolved errors while limiting incorrect non-recovery predictions.

\textbf{ReCoF maintains stable prediction performance around the chosen configuration.} Figure~\ref{fig:rq3-sensitivity} shows that small threshold offsets of $\pm0.01$ change mean Macro-F1 by at most approximately 1.43 percentage points. Using five or ten task folds also yields competitive results across the three Execution LLMs. These results indicate that ReCoF's performance does not depend on a single precise parameter setting. Performance declines under weaker regularization, particularly for Sonnet 4.5, suggesting that constraining the fusion weights helps limit overfitting. Larger threshold offsets also reduce performance, highlighting the value of validation-based calibration.

\textbf{Deciding when to intervene requires considering whether the agent can self-recover.} Our results show that this distinction can be assessed from the execution context available when an error occurs. By improving self-recovery prediction across Execution LLMs, ReCoF provides a basis for allowing agents likely to self-recover to continue and directing external assistance toward errors likely to remain unresolved.

%% file: sections/rq3-sensitivity-figure.tex
\begin{figure}[t]
\centering
\includegraphics[width=\columnwidth]{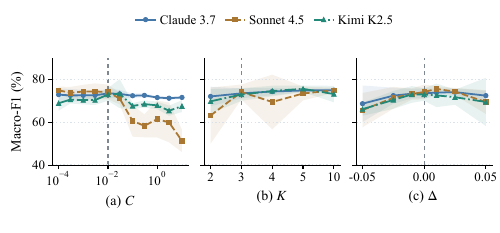}
\caption{Parameter sensitivity.}
\Description{Three side-by-side single-column panels show test Macro-F1 for Claude 3.7 Sonnet, Claude Sonnet 4.5, and Kimi K2.5. Panel (a) varies the inverse L2 regularization strength C, panel (b) the task-fold count K, and panel (c) an absolute offset from the validation-calibrated threshold. Curves are five-seed means and shaded bands are one sample standard deviation. Dashed vertical lines mark C=0.01, K=3, and zero threshold offset. Other parameters, Recovery Pattern Experts, Assessment LLM outputs and original task partitions remain fixed.}
\label{fig:rq3-sensitivity}
\end{figure}

%% file: sections/related-work.tex
\subsection{Self-Recovery in LLM Agents}
\label{sec:related-errors-recovery}

Prior work studies recovery through model training and controlled evaluation. GUI-Reflection trains GUI agents for self-reflection and error correction \citep{wu2025guireflection}. GUI-RobustEval and RoTS provide executable recovery test cases and synthesized trajectories to evaluate and improve recovery capabilities \citep{bu2026recovering}. Our study examines naturally occurring self-recovery during agents' original execution. ReInBench records error occurrence, observable recognition, and recovery, enabling analysis of how often and how quickly agents self-recover and how recovery performance varies across Execution LLMs.

Experience-based methods investigate how earlier execution influences later behavior. Reflexion uses stored reflections to guide subsequent trials \citep{shinn2023reflexion}, while Xiong et al.\ examine how memory management affects experience reuse \citep{xiong2026memory}. Our analysis examines recovery experience arising during the agent's continued execution within the same trace. We compare subsequent recovery outcomes and durations for same-type and different-type errors and examine how local self-recovery relates to overall task completion.

\subsection{Errors and Intervention}
\label{sec:related-monitoring-intervention}

Existing monitoring methods identify problematic execution or anticipate failure. AgentForesight identifies decisive errors associated with eventual failure \citep{zhang2026agentforesight}, while PrefixGuard learns execution representations for failure prediction and diagnosis \citep{huang2026prefixguard}. Their original targets concern decisive errors or task failure. Our target is whether a particular local error will subsequently self-recover.

Recovery systems additionally study how to assist agents. Wink monitors coding-agent misbehaviors and supplies corrective guidance \citep{nanda2026wink}. Sentry diagnoses failures and retrieves verified recovery lessons to guide intervention \citep{ji2026sentry}. These systems also acknowledge self-recovery through observations of redundant guidance or comparisons with unaided continuations. Xiao et al.\ compare execution with and without observation refresh and learn when this intervention improves task outcomes \citep{xiao2026harnesses}. Their decision concerns the expected benefit of a specific recovery operation.

Our study focuses on \emph{when to intervene} by examining whether an error will self-recover through continued execution. Self-recovery is relevant to this choice because an observed error may be resolved without additional assistance. We systematically analyze this behavior and adapt existing monitoring and diagnosis methods to self-recovery prediction. We further propose ReCoF to combine learned recovery patterns with contextual assessment, providing a basis for deciding whether to allow continued execution or consider external intervention.

%% file: appendices/recof-method.tex
\section{Detailed Description of ReCoF}
\label{app:recof-method}

ReCoF predicts whether an error will self-recover to inform whether additional intervention is needed. \emph{Recovery Pattern Experts} learn the relationship between historical execution contexts and recovery outcomes. \emph{Conditional Decision Fusion} relates these predictions to the Assessment LLM's contextual verdict and the Execution LLM's identity. This appendix formalizes both components and their training procedure.

\subsection{Prediction Task and Available Context}
\label{app:recof-input}

Let $e$ denote an error occurring at step $t_e$ in trace $\tau$, and let $m_e$ denote its Execution LLM. Following the notation in Section~\ref{sec:background-agents}, $g$, $r_t$, $a_t$, and $o_t$ denote the task goal, recorded reasoning, action, and resulting observation. ReCoF receives the recent execution context
\begin{equation}
    \mathcal C_e=\bigl(g,(r_s,a_s,o_s)_{s=\ell_e}^{t_e}\bigr),
    \qquad \ell_e=\max(1,t_e-H+1),
    \label{eq:recof-context}
\end{equation}
where $H=5$. Observations include available feedback and screenshots. The final step includes the recorded observation after the action at error occurrence. The target error step is supplied. Retrospective error descriptions and execution after $t_e$ are excluded. The prediction target is
\begin{equation}
    y_e=\begin{cases}
        0, & \text{if $e$ self-recovers within the observed trace},\\
        1, & \text{otherwise}.
    \end{cases}
    \label{eq:recof-target}
\end{equation}
Thus, non-recovery is the positive class. This target concerns the local error rather than final task completion. A common observation boundary makes historical training examples and prediction-time inputs comparable. Historical recovery outcomes supply the supervision. The predictor must infer each outcome from information available at error occurrence.

\subsection{Recovery Pattern Experts}
\label{app:recof-experts}

The agent's interpretation of the task, the visible application state, and recent changes in execution provide evidence about recoverability. Recovery Pattern Experts use multiple feature views and classifiers to learn how this evidence relates to recovery outcomes.

\paragraph{Execution representations.}
Frozen encoders provide text, visual, and joint image--text features. Text features represent the task, the ordered execution text, the error-occurrence step, and recorded reasoning. Visual features represent screenshot content and coarse spatial regions. Joint image--text features encode the task, Execution LLM identity, and the interleaved reasoning, actions, and available screenshots in the five-step context. The encoders are Qwen3-Embedding-0.6B, SigLIP2, and Qwen3-VL-Embedding-2B, respectively. Screenshots enter the visual and joint encoders directly. SigLIP2 patch features are spatially reduced and cached. Applying its pretrained pooling head to these cached features provides an additional visual view.

Recent changes complement the context as a whole. Let $\mathbf h_s$ denote a reasoning embedding and $\mathbf v_s$ a pooled screenshot feature. The corresponding summary features include
\begin{equation}
    \begin{aligned}
        \mathbf z_e^{\mathrm{reason}}
        &=\left[\mathbf h_{t_e},\overline{\mathbf h}_e,
            \mathbf h_{t_e}-\mathbf h_{\ell_e}\right],\\
        \mathbf z_e^{\mathrm{visual}}
        &=\left[\mathbf v_{t_e},\overline{\mathbf v}_e,
            \mathbf v_{t_e}-\mathbf v_{t_e-1}\right],
    \end{aligned}
    \label{eq:recof-temporal-features}
\end{equation}
where the bars denote window means and brackets with commas denote vector concatenation. Missing features are represented by zero vectors with availability indicators in the auxiliary descriptors. A one-step context has a zero temporal difference. Additional descriptors summarize action types and parameters, repeated actions, and textual cues. These features characterize the current execution state and recent changes.

\paragraph{Expert predictions.}
An expert selects a feature view $\mathbf z_e^{(j)}$ and produces a non-recovery score $p_{e,j}=E_j(\mathbf z_e^{(j)},m_e)$. The expert bank includes multi-view, pretrained-visual, joint image--text, joint MLP, partially pooled, and robust RBF predictors, together with an ensemble derived from validation-subset voting. For the multi-view, pretrained-visual, and joint image--text predictors, we fit candidate classifiers on the corresponding feature representations and select each predictor using validation performance. The joint MLP uses a compact nonlinear head, whereas RBF predictors compare nonlinear neighborhoods in the encoded feature space. Classifier parameters are fitted on training events, and configurations are selected using validation events. Outputs are mapped to $[0,1]$. For an SVM, this mapping applies a sigmoid to its decision score.

Most expert families use model-specific predictors. The partially pooled family also shares information across Execution LLMs through the kernel
\begin{equation}
    k\bigl((\mathbf z,m),(\mathbf z',m')\bigr)
    =\exp\!\left(-\gamma\|\mathbf z-\mathbf z'\|^2\right)
      \frac{1+\eta\,\mathbb I[m=m']}{1+\eta},
    \label{eq:recof-pooled-kernel}
\end{equation}
where $\eta\geq0$ controls the relative contribution of examples from the same Execution LLM. When $\eta=0$, the kernel depends only on feature similarity. Increasing $\eta$ reduces the weight assigned to examples from other Execution LLMs. Prediction scores are collected in the vector
\begin{equation}
    \mathbf p_e=\bigl[p_{e,1},\ldots,p_{e,D}\bigr]^{\mathsf T}
    \in[0,1]^D,
    \label{eq:recof-expert-scores}
\end{equation}
where $D$ denotes the number of prediction score channels. The robust RBF predictor is selected by the lower decile of bootstrap validation AUC. The ensemble entry uses the first five selected predictor families. In each of 200 label-stratified validation subsets containing 70\% of each class, it selects a predictor and threshold by Macro-F1. Its test score is the fraction of non-recovery votes. For a validation event, only votes from subsets excluding that event contribute to its ensemble score.

This design connects historical recovery outcomes to observable execution patterns. Separate text and visual views retain evidence about the agent's interpretation and the application state. The joint view captures their relationship. Temporal summaries indicate how that evidence changes near the error. Different classifier families provide alternative decision boundaries over these representations. Model-specific predictors preserve differences in recovery capability, while partial pooling permits shared patterns across Execution LLMs. Freezing the encoders concentrates supervised learning on the prediction layers, limiting the number of parameters learned from the annotated events.

\subsection{Conditional Decision Fusion}
\label{app:recof-fusion}

Similar execution patterns can have different recovery outcomes under different task requirements. Conditional Decision Fusion uses contextual assessment to adjust expert contributions to the current error context.

\paragraph{Contextual verdict.}
The Assessment LLM receives the task goal and five-step execution text. It also receives the screenshot at error occurrence, when recorded, and the most recent earlier screenshot in the same window. Its instruction distinguishes correcting the target error from completing the task and asks whether the error will self-recover. The Assessment LLM returns a binary verdict $a_e$, with $a_e=1$ for predicted non-recovery and $a_e=0$ for predicted self-recovery. ReCoF uses this verdict to condition the contribution of the expert predictions.

\paragraph{Conditional interaction.}
Let $\mathbf u_{m_e}\in\{0,1\}^{M}$ be the one-hot identity of the Execution LLM, with $M=3$ in our experiments. We construct
\begin{equation}
    \begin{aligned}
        \mathbf r_e&=\mathbf p_e\odot(2a_e-1),\\
        \mathbf x_e&=\left[\mathbf p_e,a_e,\mathbf r_e,\mathbf u_{m_e}\right],
    \end{aligned}
    \label{eq:recof-interaction}
\end{equation}
where $\odot$ denotes element-wise multiplication by the broadcast scalar. The fusion vector $\mathbf x_e$ has $2D+M+1$ dimensions. For fusion fold $k$, the non-recovery probability is
\begin{equation}
    \begin{aligned}
    q_e^{(k)}&=\sigma\!\left(b_k+\mathbf w_k^{\mathsf T}
        S_k(\mathbf x_e)\right),\\
    S_k(\mathbf x)&=\operatorname{clip}\!\left(
        \frac{\mathbf x-\boldsymbol\mu_k}{\mathbf d_k},-8,8\right),
    \end{aligned}
    \label{eq:recof-logistic}
\end{equation}
where $\sigma$ is the sigmoid, and $\boldsymbol\mu_k$ and $\mathbf d_k$ are the feature means and standard deviations fitted on that fold's training rows. Zero-variance coordinates use a scale of one.

The interaction changes how the verdict affects expert evidence. In the unclipped region, after absorbing affine scaling into the coefficients, the contribution of expert $j$ has the effective coefficient
\begin{equation}
    \widetilde w_j(a_e)=w_j^{(p)}+(2a_e-1)w_j^{(r)}.
    \label{eq:recof-effective-weight}
\end{equation}
Consequently, the same expert score can receive different contributions under predicted self-recovery and predicted non-recovery. Without the interaction, the verdict changes only an additive term in the logit. Execution LLM identity contributes a model-specific additive adjustment. Model-specific experts and decision thresholds provide the remaining adaptation to the Execution LLM.

This design lets the fusion learn when to reinforce or discount historical predictions according to the contextual verdict. The direct contribution of the Assessment LLM and its interaction with each expert are learned from recovery outcomes. By conditioning expert contributions on the contextual verdict and incorporating the Execution LLM's identity, ReCoF accounts for both the current error context and differences in recovery capability.

\subsection{Fusion Training and Decision Calibration}
\label{app:recof-training}

The base experts are fixed before fusion training. We divide validation tasks into $K=3$ folds, assigning every event from the same task to the same fold. Each fusion model is fitted on the other $K-1$ folds. For its training set $\mathcal V_{-k}$, let $N_k=|\mathcal V_{-k}|$ and let $n_{k,m,y}$ count examples from Execution LLM $m$ with label $y$. We use weights $\omega_e=N_k/(2M n_{k,m_e,y_e})$, giving equal total weight to each model--class group. The regularized logistic objective is
\begin{equation}
    \begin{aligned}
    \min_{\mathbf w_k,b_k}\quad&
        \frac{\|\mathbf w_k\|_2^2+b_k^2}{2}
        +C\sum_{e\in\mathcal V_{-k}}\omega_e\,
            \ell\bigl(y_e,q_e^{(k)}\bigr),\\
    \ell(y,q)&=-y\log q-(1-y)\log(1-q),
    \end{aligned}
    \label{eq:recof-fusion-objective}
\end{equation}
with $C=0.01$. Grouping by task prevents related events from appearing in both sides of a fusion fold. Balancing model--class groups prevents the largest execution-model or outcome group from dominating the objective. Regularization limits the complexity of the conditional score combinations.

Each validation event receives an out-of-fold probability from the fusion model that excludes its task. For each Execution LLM $m$, we draw $B=200$ label-stratified bootstrap samples from these validation predictions. If $\theta_m^{(b)}$ maximizes Macro-F1 in draw $b$, the final threshold is $\theta_m=\operatorname{median}_{b=1}^{B}\theta_m^{(b)}$. Candidate thresholds are the midpoints between consecutive distinct validation scores, two boundary candidates, and $0.5$. Ties favor the candidate closest to $0.5$, then the smaller value. At prediction time,
\begin{equation}
    \overline q_e=\frac{1}{K}\sum_{k=1}^{K}q_e^{(k)},
    \qquad \widehat y_e=\mathbb I[\overline q_e\geq\theta_{m_e}].
    \label{eq:recof-final-prediction}
\end{equation}
The fold models are retained without a final refit. Averaging reduces dependence on one validation partition, while model-specific thresholds accommodate differences in decision calibration. The median of bootstrap optima reduces dependence on a single small validation sample. The validation partition is used for expert selection and fusion calibration. The task-grouped folds provide out-of-fold predictions for the fusion layer, and final performance is evaluated on the task-disjoint test partition.

ReCoF uses the execution context available at error occurrence to distinguish errors likely to self-recover from those likely to remain unresolved. These predictions provide a basis for deciding when to allow continued execution and when to prioritize external intervention.

%% file: main.bbl

\begin{thebibliography}{22}


\ifx \showCODEN    \undefined \def \showCODEN     #1{\unskip}     \fi
\ifx \showISBNx    \undefined \def \showISBNx     #1{\unskip}     \fi
\ifx \showISBNxiii \undefined \def \showISBNxiii  #1{\unskip}     \fi
\ifx \showISSN     \undefined \def \showISSN      #1{\unskip}     \fi
\ifx \showLCCN     \undefined \def \showLCCN      #1{\unskip}     \fi
\ifx \shownote     \undefined \def \shownote      #1{#1}          \fi
\ifx \showarticletitle \undefined \def \showarticletitle #1{#1}   \fi
\ifx \showURL      \undefined \def \showURL       {\relax}        \fi
\providecommand\bibfield[2]{#2}
\providecommand\bibinfo[2]{#2}
\providecommand\natexlab[1]{#1}
\providecommand\showeprint[2][]{arXiv:#2}

\bibitem[Avi{\v z}ienis et~al\mbox{.}(2004)]%
        {avizienis2004basic}
\bibfield{author}{\bibinfo{person}{Algirdas Avi{\v z}ienis},
  \bibinfo{person}{Jean-Claude Laprie}, \bibinfo{person}{Brian Randell}, {and}
  \bibinfo{person}{Carl Landwehr}.} \bibinfo{year}{2004}\natexlab{}.
\newblock \showarticletitle{Basic Concepts and Taxonomy of Dependable and
  Secure Computing}.
\newblock \bibinfo{journal}{\emph{IEEE Transactions on Dependable and Secure
  Computing}} \bibinfo{volume}{1}, \bibinfo{number}{1} (\bibinfo{year}{2004}),
  \bibinfo{pages}{11--33}.
\newblock
\href{https://doi.org/10.1109/TDSC.2004.2}{doi:\nolinkurl{10.1109/TDSC.2004.2}}


\bibitem[Barbi et~al\mbox{.}(2025)]%
        {barbi2025preventing}
\bibfield{author}{\bibinfo{person}{Ohav Barbi}, \bibinfo{person}{Ori Yoran},
  {and} \bibinfo{person}{Mor Geva}.} \bibinfo{year}{2025}\natexlab{}.
\newblock \showarticletitle{Preventing Rogue Agents Improves Multi-Agent
  Collaboration}. In \bibinfo{booktitle}{\emph{Proceedings of the 1st Workshop
  for Research on Agent Language Models (REALM 2025)}}.
  \bibinfo{publisher}{Association for Computational Linguistics},
  \bibinfo{address}{Vienna, Austria}, \bibinfo{pages}{486--511}.
\newblock
\urldef\tempurl%
\url{https://aclanthology.org/2025.realm-1.34/}
\showURL{%
\tempurl}


\bibitem[Bu et~al\mbox{.}(2026)]%
        {bu2026recovering}
\bibfield{author}{\bibinfo{person}{Tianpeng Bu}, \bibinfo{person}{Xin Liu},
  \bibinfo{person}{Qihua Chen}, \bibinfo{person}{Hao Jiang},
  \bibinfo{person}{Shurui Li}, \bibinfo{person}{Hongtao Duan},
  \bibinfo{person}{Lu Jiang}, \bibinfo{person}{Lulu Hu}, \bibinfo{person}{Bin
  Yang}, {and} \bibinfo{person}{Minying Zhang}.}
  \bibinfo{year}{2026}\natexlab{}.
\newblock \showarticletitle{Recovering Policy-Induced Errors: Benchmarking and
  Trajectory Synthesis for Robust {GUI} Agents}. In
  \bibinfo{booktitle}{\emph{Proceedings of the 43rd International Conference on
  Machine Learning}} \emph{(\bibinfo{series}{Proceedings of Machine Learning
  Research}, Vol.~\bibinfo{volume}{306})}. \bibinfo{publisher}{PMLR},
  \bibinfo{pages}{10111--10146}.
\newblock
\urldef\tempurl%
\url{https://proceedings.mlr.press/v306/bu26b.html}
\showURL{%
\tempurl}


\bibitem[Cai et~al\mbox{.}(2025)]%
        {cai2025personalized}
\bibfield{author}{\bibinfo{person}{Hongru Cai}, \bibinfo{person}{Yongqi Li},
  \bibinfo{person}{Wenjie Wang}, \bibinfo{person}{Fengbin Zhu},
  \bibinfo{person}{Xiaoyu Shen}, \bibinfo{person}{Wenjie Li}, {and}
  \bibinfo{person}{Tat-Seng Chua}.} \bibinfo{year}{2025}\natexlab{}.
\newblock \showarticletitle{Large Language Models Empowered Personalized Web
  Agents}. In \bibinfo{booktitle}{\emph{Proceedings of the ACM on Web
  Conference 2025}}. \bibinfo{publisher}{Association for Computing Machinery},
  \bibinfo{address}{New York, NY, USA}, \bibinfo{pages}{198--215}.
\newblock
\href{https://doi.org/10.1145/3696410.3714842}{doi:\nolinkurl{10.1145/3696410.3714842}}


\bibitem[Deng et~al\mbox{.}(2023)]%
        {deng2023mind2web}
\bibfield{author}{\bibinfo{person}{Xiang Deng}, \bibinfo{person}{Yu Gu},
  \bibinfo{person}{Boyuan Zheng}, \bibinfo{person}{Shijie Chen},
  \bibinfo{person}{Samuel Stevens}, \bibinfo{person}{Boshi Wang},
  \bibinfo{person}{Huan Sun}, {and} \bibinfo{person}{Yu Su}.}
  \bibinfo{year}{2023}\natexlab{}.
\newblock \showarticletitle{{Mind2Web}: Towards a Generalist Agent for the
  Web}. In \bibinfo{booktitle}{\emph{Advances in Neural Information Processing
  Systems}}, Vol.~\bibinfo{volume}{36}. \bibinfo{publisher}{Curran Associates,
  Inc.}, \bibinfo{pages}{28091--28114}.
\newblock
\urldef\tempurl%
\url{https://proceedings.neurips.cc/paper_files/paper/2023/file/5950bf290a1570ea401bf98882128160-Paper-Datasets_and_Benchmarks.pdf}
\showURL{%
\tempurl}


\bibitem[{HAL}(2025)]%
        {hal2025seeact}
\bibfield{author}{\bibinfo{person}{{HAL}}.} \bibinfo{year}{2025}\natexlab{}.
\newblock \bibinfo{title}{{SeeAct}: {Online-Mind2Web} Agent Performance and
  Archived Traces}.
\newblock \bibinfo{howpublished}{Holistic Agent Leaderboard}.
\newblock
\urldef\tempurl%
\url{https://hal.cs.princeton.edu/agent/SeeAct}
\showURL{%
\tempurl}
\newblock
\shownote{GPT-5 medium run dated August 13, 2025; accessed October 18, 2026}.


\bibitem[Huang et~al\mbox{.}(2026)]%
        {huang2026prefixguard}
\bibfield{author}{\bibinfo{person}{Xinmiao Huang}, \bibinfo{person}{Jinwei Hu},
  \bibinfo{person}{Qisong He}, \bibinfo{person}{Rajarshi Roy},
  \bibinfo{person}{Changshun Wu}, \bibinfo{person}{Yi Dong}, {and}
  \bibinfo{person}{Xiaowei Huang}.} \bibinfo{year}{2026}\natexlab{}.
\newblock \showarticletitle{{PrefixGuard}: Online Failure Warning and
  Trace-Grounded Diagnosis for {LLM} Agents}.
\newblock \bibinfo{journal}{\emph{arXiv preprint arXiv:2605.06455}}
  (\bibinfo{year}{2026}).
\newblock
\urldef\tempurl%
\url{https://arxiv.org/abs/2605.06455v2}
\showURL{%
\tempurl}
\newblock
\shownote{Version 2, September 27, 2026}.


\bibitem[Ji et~al\mbox{.}(2026)]%
        {ji2026sentry}
\bibfield{author}{\bibinfo{person}{Changxiu Ji}, \bibinfo{person}{Amy Lu},
  \bibinfo{person}{Qizheng Zhang}, {and} \bibinfo{person}{Kunle Olukotun}.}
  \bibinfo{year}{2026}\natexlab{}.
\newblock \showarticletitle{Sentry: Learning to Recover from {LLM} Agent
  Failures at Test Time}.
\newblock \bibinfo{journal}{\emph{arXiv preprint arXiv:2610.02994}}
  (\bibinfo{year}{2026}).
\newblock
\urldef\tempurl%
\url{https://arxiv.org/abs/2610.02994v1}
\showURL{%
\tempurl}
\newblock
\shownote{Version 1, October 2, 2026}.


\bibitem[Li et~al\mbox{.}(2023)]%
        {li2023coannotating}
\bibfield{author}{\bibinfo{person}{Minzhi Li}, \bibinfo{person}{Taiwei Shi},
  \bibinfo{person}{Caleb Ziems}, \bibinfo{person}{Min-Yen Kan},
  \bibinfo{person}{Nancy~F. Chen}, \bibinfo{person}{Zhengyuan Liu}, {and}
  \bibinfo{person}{Diyi Yang}.} \bibinfo{year}{2023}\natexlab{}.
\newblock \showarticletitle{{CoAnnotating}: Uncertainty-Guided Work Allocation
  between Human and Large Language Models for Data Annotation}. In
  \bibinfo{booktitle}{\emph{Proceedings of the 2023 Conference on Empirical
  Methods in Natural Language Processing}}. \bibinfo{publisher}{Association for
  Computational Linguistics}, \bibinfo{address}{Singapore},
  \bibinfo{pages}{1487--1505}.
\newblock
\href{https://doi.org/10.18653/v1/2023.emnlp-main.92}{doi:\nolinkurl{10.18653/v1/2023.emnlp-main.92}}


\bibitem[Nanda et~al\mbox{.}(2026)]%
        {nanda2026wink}
\bibfield{author}{\bibinfo{person}{Rahul Nanda}, \bibinfo{person}{Chandra
  Maddila}, \bibinfo{person}{Smriti Jha}, \bibinfo{person}{Euna~Mehnaz Khan},
  \bibinfo{person}{Satish Chandra}, {and} \bibinfo{person}{Matteo Paltenghi}.}
  \bibinfo{year}{2026}\natexlab{}.
\newblock \showarticletitle{Wink: Recovering from Misbehaviors in Coding
  Agents}. In \bibinfo{booktitle}{\emph{Proceedings of the 3rd ACM
  International Conference on AI-powered Software}}
  \emph{(\bibinfo{series}{AIware '26})}. \bibinfo{publisher}{Association for
  Computing Machinery}, \bibinfo{address}{New York, NY, USA}.
\newblock
\href{https://doi.org/10.1145/3805760.3814911}{doi:\nolinkurl{10.1145/3805760.3814911}}


\bibitem[Shi et~al\mbox{.}(2025)]%
        {shi2025toolwild}
\bibfield{author}{\bibinfo{person}{Zhengliang Shi}, \bibinfo{person}{Shen Gao},
  \bibinfo{person}{Lingyong Yan}, \bibinfo{person}{Yue Feng},
  \bibinfo{person}{Xiuyi Chen}, \bibinfo{person}{Zhumin Chen},
  \bibinfo{person}{Dawei Yin}, \bibinfo{person}{Suzan Verberne}, {and}
  \bibinfo{person}{Zhaochun Ren}.} \bibinfo{year}{2025}\natexlab{}.
\newblock \showarticletitle{Tool Learning in the Wild: Empowering Language
  Models as Automatic Tool Agents}. In \bibinfo{booktitle}{\emph{Proceedings of
  the ACM on Web Conference 2025}}. \bibinfo{publisher}{Association for
  Computing Machinery}, \bibinfo{address}{New York, NY, USA},
  \bibinfo{pages}{2222--2237}.
\newblock
\href{https://doi.org/10.1145/3696410.3714825}{doi:\nolinkurl{10.1145/3696410.3714825}}


\bibitem[Shinn et~al\mbox{.}(2023)]%
        {shinn2023reflexion}
\bibfield{author}{\bibinfo{person}{Noah Shinn}, \bibinfo{person}{Federico
  Cassano}, \bibinfo{person}{Ashwin Gopinath}, \bibinfo{person}{Karthik
  Narasimhan}, {and} \bibinfo{person}{Shunyu Yao}.}
  \bibinfo{year}{2023}\natexlab{}.
\newblock \showarticletitle{Reflexion: Language Agents with Verbal
  Reinforcement Learning}. In \bibinfo{booktitle}{\emph{Advances in Neural
  Information Processing Systems}}, Vol.~\bibinfo{volume}{36}.
\newblock
\urldef\tempurl%
\url{https://proceedings.neurips.cc/paper_files/paper/2023/file/1b44b878bb782e6954cd888628510e90-Paper-Conference.pdf}
\showURL{%
\tempurl}


\bibitem[Wang et~al\mbox{.}(2024)]%
        {wang2024collaborative}
\bibfield{author}{\bibinfo{person}{Xinru Wang}, \bibinfo{person}{Hannah Kim},
  \bibinfo{person}{Sajjadur Rahman}, \bibinfo{person}{Kushan Mitra}, {and}
  \bibinfo{person}{Zhengjie Miao}.} \bibinfo{year}{2024}\natexlab{}.
\newblock \showarticletitle{Human-{LLM} Collaborative Annotation Through
  Effective Verification of {LLM} Labels}. In
  \bibinfo{booktitle}{\emph{Proceedings of the CHI Conference on Human Factors
  in Computing Systems}}. \bibinfo{publisher}{Association for Computing
  Machinery}, \bibinfo{address}{New York, NY, USA}, Article
  \bibinfo{articleno}{303}, \bibinfo{numpages}{21}~pages.
\newblock
\href{https://doi.org/10.1145/3613904.3641960}{doi:\nolinkurl{10.1145/3613904.3641960}}


\bibitem[Wu et~al\mbox{.}(2025)]%
        {wu2025guireflection}
\bibfield{author}{\bibinfo{person}{Penghao Wu}, \bibinfo{person}{Shengnan Ma},
  \bibinfo{person}{Bo Wang}, \bibinfo{person}{Jiaheng Yu},
  \bibinfo{person}{Lewei Lu}, {and} \bibinfo{person}{Ziwei Liu}.}
  \bibinfo{year}{2025}\natexlab{}.
\newblock \showarticletitle{{GUI-Reflection}: Empowering Multimodal {GUI}
  Models with Self-Reflection Behavior}. In \bibinfo{booktitle}{\emph{Advances
  in Neural Information Processing Systems}}, Vol.~\bibinfo{volume}{38}.
\newblock
\urldef\tempurl%
\url{https://papers.nips.cc/paper_files/paper/2025/hash/937ae0e83eb08d2cb8627fe1def8c751-Abstract-Conference.html}
\showURL{%
\tempurl}


\bibitem[Xiao et~al\mbox{.}(2026)]%
        {xiao2026harnesses}
\bibfield{author}{\bibinfo{person}{Shuyao Xiao}, \bibinfo{person}{Shengling
  Wang}, \bibinfo{person}{Xuan Chen}, \bibinfo{person}{Ke Chao},
  \bibinfo{person}{Ming Cui}, \bibinfo{person}{Feifei Qian},
  \bibinfo{person}{Chaoyang Mei}, \bibinfo{person}{Fanlin Meng},
  \bibinfo{person}{Ziming Yu}, {and} \bibinfo{person}{Junxi Yin}.}
  \bibinfo{year}{2026}\natexlab{}.
\newblock \showarticletitle{When Harnesses Lose the Signal: Causal Evaluation
  of Recovery in {LLM} Agents}.
\newblock \bibinfo{journal}{\emph{arXiv preprint arXiv:2610.00372}}
  (\bibinfo{year}{2026}).
\newblock
\urldef\tempurl%
\url{https://arxiv.org/abs/2610.00372v1}
\showURL{%
\tempurl}
\newblock
\shownote{Version 1}.


\bibitem[Xie et~al\mbox{.}(2024)]%
        {xie2024osworld}
\bibfield{author}{\bibinfo{person}{Tianbao Xie}, \bibinfo{person}{Danyang
  Zhang}, \bibinfo{person}{Jixuan Chen}, \bibinfo{person}{Xiaochuan Li},
  \bibinfo{person}{Siheng Zhao}, \bibinfo{person}{Ruisheng Cao},
  \bibinfo{person}{Toh~Jing Hua}, \bibinfo{person}{Zhoujun Cheng},
  \bibinfo{person}{Dongchan Shin}, \bibinfo{person}{Fangyu Lei},
  \bibinfo{person}{Yitao Liu}, \bibinfo{person}{Yiheng Xu},
  \bibinfo{person}{Shuyan Zhou}, \bibinfo{person}{Silvio Savarese},
  \bibinfo{person}{Caiming Xiong}, \bibinfo{person}{Victor Zhong}, {and}
  \bibinfo{person}{Tao Yu}.} \bibinfo{year}{2024}\natexlab{}.
\newblock \showarticletitle{{OSWorld}: Benchmarking Multimodal Agents for
  Open-Ended Tasks in Real Computer Environments}. In
  \bibinfo{booktitle}{\emph{Advances in Neural Information Processing
  Systems}}, Vol.~\bibinfo{volume}{37}.
\newblock
\urldef\tempurl%
\url{https://proceedings.neurips.cc/paper_files/paper/2024/hash/5d413e48f84dc61244b6be550f1cd8f5-Abstract-Datasets_and_Benchmarks_Track.html}
\showURL{%
\tempurl}


\bibitem[Xiong et~al\mbox{.}(2026)]%
        {xiong2026memory}
\bibfield{author}{\bibinfo{person}{Zidi Xiong}, \bibinfo{person}{Yuping Lin},
  \bibinfo{person}{Wenya Xie}, \bibinfo{person}{Pengfei He},
  \bibinfo{person}{Zirui Liu}, \bibinfo{person}{Jiliang Tang},
  \bibinfo{person}{Himabindu Lakkaraju}, {and} \bibinfo{person}{Zhen Xiang}.}
  \bibinfo{year}{2026}\natexlab{}.
\newblock \showarticletitle{How Memory Management Impacts {LLM} Agents: An
  Empirical Study of Experience-Following Behavior}. In
  \bibinfo{booktitle}{\emph{Proceedings of the 64th Annual Meeting of the
  Association for Computational Linguistics (Volume 1: Long Papers)}}.
  \bibinfo{publisher}{Association for Computational Linguistics},
  \bibinfo{pages}{623--645}.
\newblock
\href{https://doi.org/10.18653/v1/2026.acl-long.27}{doi:\nolinkurl{10.18653/v1/2026.acl-long.27}}


\bibitem[{xlangai}(nd)]%
        {osworldverifiedtraces}
\bibfield{author}{\bibinfo{person}{{xlangai}}.}
  \bibinfo{year}{n.d.}\natexlab{}.
\newblock \bibinfo{title}{{OSWorld-Verified} Model Trajectories}.
\newblock \bibinfo{howpublished}{Hugging Face dataset}.
\newblock
\urldef\tempurl%
\url{https://huggingface.co/datasets/xlangai/ubuntu_osworld_verified_trajs}
\showURL{%
\tempurl}
\newblock
\shownote{Accessed October 15, 2026}.


\bibitem[Xue et~al\mbox{.}(2025)]%
        {xue2025online}
\bibfield{author}{\bibinfo{person}{Tianci Xue}, \bibinfo{person}{Weijian Qi},
  \bibinfo{person}{Tianneng Shi}, \bibinfo{person}{Chan~Hee Song},
  \bibinfo{person}{Boyu Gou}, \bibinfo{person}{Dawn Song},
  \bibinfo{person}{Huan Sun}, {and} \bibinfo{person}{Yu Su}.}
  \bibinfo{year}{2025}\natexlab{}.
\newblock \showarticletitle{An Illusion of Progress? Assessing the Current
  State of Web Agents}. In \bibinfo{booktitle}{\emph{Second Conference on
  Language Modeling}}.
\newblock
\urldef\tempurl%
\url{https://openreview.net/forum?id=6jZi4HSs6o}
\showURL{%
\tempurl}


\bibitem[Yao et~al\mbox{.}(2023)]%
        {yao2023react}
\bibfield{author}{\bibinfo{person}{Shunyu Yao}, \bibinfo{person}{Jeffrey Zhao},
  \bibinfo{person}{Dian Yu}, \bibinfo{person}{Nan Du}, \bibinfo{person}{Izhak
  Shafran}, \bibinfo{person}{Karthik Narasimhan}, {and} \bibinfo{person}{Yuan
  Cao}.} \bibinfo{year}{2023}\natexlab{}.
\newblock \showarticletitle{{ReAct}: Synergizing Reasoning and Acting in
  Language Models}. In \bibinfo{booktitle}{\emph{International Conference on
  Learning Representations}}.
\newblock
\urldef\tempurl%
\url{https://arxiv.org/abs/2210.03629}
\showURL{%
\tempurl}


\bibitem[Zhang et~al\mbox{.}(2026)]%
        {zhang2026agentforesight}
\bibfield{author}{\bibinfo{person}{Boxuan Zhang}, \bibinfo{person}{Jianing
  Zhu}, \bibinfo{person}{Zeru Shi}, \bibinfo{person}{Dongfang Liu}, {and}
  \bibinfo{person}{Ruixiang Tang}.} \bibinfo{year}{2026}\natexlab{}.
\newblock \showarticletitle{{AgentForesight}: Online Auditing for Early Failure
  Prediction in Multi-Agent Systems}.
\newblock \bibinfo{journal}{\emph{arXiv preprint arXiv:2605.08715}}
  (\bibinfo{year}{2026}).
\newblock
\urldef\tempurl%
\url{https://arxiv.org/abs/2605.08715v1}
\showURL{%
\tempurl}
\newblock
\shownote{Version 1, May 9, 2026}.


\bibitem[Zhou et~al\mbox{.}(2024)]%
        {zhou2024webarena}
\bibfield{author}{\bibinfo{person}{Shuyan Zhou}, \bibinfo{person}{Frank~F. Xu},
  \bibinfo{person}{Hao Zhu}, \bibinfo{person}{Xuhui Zhou},
  \bibinfo{person}{Robert Lo}, \bibinfo{person}{Abishek Sridhar},
  \bibinfo{person}{Xianyi Cheng}, \bibinfo{person}{Tianyue Ou},
  \bibinfo{person}{Yonatan Bisk}, \bibinfo{person}{Daniel Fried},
  \bibinfo{person}{Uri Alon}, {and} \bibinfo{person}{Graham Neubig}.}
  \bibinfo{year}{2024}\natexlab{}.
\newblock \showarticletitle{{WebArena}: A Realistic Web Environment for
  Building Autonomous Agents}. In \bibinfo{booktitle}{\emph{International
  Conference on Learning Representations}}.
\newblock
\urldef\tempurl%
\url{https://openreview.net/forum?id=oKn9c6ytLx}
\showURL{%
\tempurl}


\end{thebibliography}
